\documentclass[11pt]{article}

\usepackage[final]{acl}

\usepackage{times}
\usepackage{latexsym}

\usepackage[T1]{fontenc}

\usepackage[utf8]{inputenc}

\usepackage{microtype}

\usepackage{inconsolata}

\usepackage{graphicx}

\usepackage{xspace}

\usepackage{enumitem}
\usepackage{amsmath}
\usepackage{float}
\usepackage{amssymb}
\usepackage[most]{tcolorbox}
\newtcolorbox{promptblock}{
  enhanced,
  colback=gray!3,
  colframe=black!45,
  boxrule=0.5pt,
  arc=2pt,
  left=7pt,
  right=7pt,
  top=1pt,
  bottom=1pt,
}
\usepackage[capitalize,noabbrev,nameinlink]{cleveref}
\usepackage{multirow}
\usepackage{booktabs}
\usepackage{url}
\title{Self-supervised User Profile Generation for Personalization}

\author{Clark Mingxuan Ju, Yuwei Qiu, Tong Zhao, Neil Shah \\
  Snap Inc. \\
  Bellevue, WA USA\\
  \texttt{mju,yqiu,tzhao,nshah@snap.com}}

\begin{document}
\maketitle
\begin{abstract}
Personalizing large language models (LLMs) has become a central challenge as LLMs are deployed across recommendation, search, dialogue, and content generation -- settings where the same query should yield different answers given different users.
A promising route is to summarize each user's interaction history into a natural-language memory or profile and prepend it to the prompt to facilitate personalization.
Existing methods learn such profile generators with explicit rewards derived from labeled downstream tasks, which are expensive and sparse as they require annotated supervision for every target task.
In light of this challenge, we introduce \textbf{B}idirectional \textbf{U}ser \textbf{M}odeling via \textbf{P}rofiles (\textbf{BUMP}), a self-supervised framework that trains a profile generator without any downstream labels. 
Specifically, given a user's interaction history, we use GRPO to train an LLM to emit a free-form textual profile under a bidirectional in-batch ranking objective: a small LLM judge measures (i) how well the generated profile, used as a query, ranks the user's own held-out interactions above interactions from other users in the batch, and (ii) how well a held-out interaction, used as a query, ranks the user's own profile above profiles of other users. 
Both directions are scored with multi-positive NDCG and combined into a dense reward per rollout; other users in the batch supply free negatives, so every training example yields supervision from raw interaction logs alone. 
Evaluated on the LaMP benchmark, BUMP matches or outperforms closed-source APIs and prior methods relying on labeled rewards, while requiring no task label at training. 
\end{abstract}

\section{Introduction}
Personalization has emerged as a fundamental capability for modern large language models (LLMs)~\citep{chen2024large}. 
As LLMs are deployed across recommendation~\citep{geng2022recommendation}, search~\citep{jin2025search}, dialogue~\citep{thoppilan2022lamda}, writing assistance~\citep{mysore2024pearl}, and content generation~\citep{salemi2024lamp}, the same query is expected to produce different and contextually appropriate responses for different users. 
Traditional tools for representing users (e.g., collaborative-filtering embeddings)~\citep{rendle2012bpr,ju2025learning} capture statistical regularities but do not interface naturally with instruction-tuned LLMs, whose only conditioning surface is the prompt itself ~\citep{brown2020language}. As there are increasingly more closed-weight model and agent experiences which users can only personalize via inputs, personalization through a representational view also becomes infeasible.  
This mismatch has driven a wave of interest in LLM-native natural-language user profiles that can be organically consumed by a frozen (and possibly private) backbone~\citep{wang2025lettingo,shi2025personax,salemi2024lamp}, injecting per-user signal without model retraining. Such natural-language profiles are also alluring for cross-domain and zero-shot transfer capabilities.

The simplest baseline for natural-language profiles is to concatenate a user's raw interaction history into a prompt and let the LLMs attend over it directly~\citep{salemi2024lamp}. 
In practice, this is prohibitive at deployment scale. 
Real users accumulate hundreds to thousands of interactions over months or years~\citep{ju2025learning,hou2024bridging}, and stuffing the full history into every request forces the model to prefill an extremely long context on each call, inflating latency, KV-cache memory, and per-query inference cost, and pushing many users past the model's context window entirely~\citep{richardson2023integrating,shi2025personax}. 

Even when the window nominally fits, long-context attention degrades: salient signal gets ``lost in the middle''~\citep{liu2024lost,huang2026threshold} and irrelevant interactions dilute the personalization quality. 
A natural-language profile sidesteps both failure modes: it compresses raw behavior into a compact, fixed-size representation that can be precomputed offline once per user, cached, and reused across arbitrarily many downstream queries, amortizing the prefill cost over an entire user session rather than paying it per request~\citep{richardson2023integrating,arora2026high}. 

A growing body of work moves beyond prompting and explicitly \emph{learns} the profile generator. \citet{richardson2023integrating} fine-tune an LLM to produce task-aware textual summaries of user history. 
\citet{arora2026high} train a 1.7B-parameter policy with GRPO, rewarding profiles that downstream engagement-prediction oracles can use to recover click signals. 
\citet{lin2025rec} close the loop end-to-end, treating a downstream recommender as the reward function for a profile/query generator. 
Profile-augmented variants of per-user PEFT methods~\citep{tan2024democratizing} likewise distill ``good'' profiles by reference to a labeled downstream task. All these methods share a single supervision recipe: a \emph{labeled downstream task} defines profile quality.
While effective, this recipe has severe limitations: rewards are \emph{expensive} and \emph{not scalable} -- each target task requires an annotated corpus and fresh label collection for policy retraining, resulting in expensive task-specific profiles which undercut generality and shared latent user interests.

    

In light of this challenge, we introduce \textbf{B}idirectional \textbf{U}ser \textbf{M}odeling via \textbf{P}rofiles (\textbf{BUMP}), a self-supervised framework that trains a profile generator without any downstream task labels. 
Our intuition derives from two principles: a good profile should be predictive of future behaviors, and conversely identifiable from these behaviors.  These properties  can be measured directly against the user's own held-out interactions, without external annotation. Hence, we turn this self-consistency principle into a powerful training signal. 

Specifically, given a batch of users, we ask an LLM policy to emit a free-form textual profile, or user summary, and a small frozen LLM judge scores it in two complementary directions: a \emph{forward} direction in which the generated profile is used as a query to rank the user's held-out interactions against in-batch interactions from other users, and a \emph{backward} direction in which a held-out interaction is used as a query to rank the user's profile against in-batch profiles from other users. 
Both directions are scored with multi-positive NDCG and combined into a single dense per-rollout reward that drives policy optimization with GRPO~\citep{shao2024deepseekmath}.

In BUMP, because every other user in the batch supplies free negatives, each training example yields supervision from raw interaction logs alone -- no downstream task is invoked, no annotation is collected, and the supervision budget grows automatically with the user population rather than the labeling budget. 
Self-supervision also frees the profile from any single downstream task's bias: the user profile is optimized to be a faithful description of the user and not a feature engineered for any one particular classifier, so the same profile transfers across tasks without retraining the generator. 
Our contributions are threefold:
\begin{itemize}[leftmargin=*]
    \item \textbf{A novel self-supervised objective for user profile generation.} We formulate user profile generation as a bidirectional in-batch ranking problem, deriving dense supervision from raw interaction logs without any downstream task label.
    \item \textbf{BUMP and BUMP+.} 
    We instantiate the objective as \textbf{BUMP}, a GRPO-trained profile generator scored by a frozen LLM judge with multi-positive NDCG over in-batch negatives, and propose \textbf{BUMP+}, which strengthens the signal via hard-negative mining in both directions.
    \item \textbf{Strong empirical results.} 
    On the LaMP benchmark, our proposals match or outperform closed-source API baselines packed with more parameters and prior methods trained with explicit labeled downstream rewards. 
\end{itemize}

\section{Related Work}
\subsection{LLM Personalization}
  Personalization for instruction-tuned LLMs has received substantial recent attention, summarized in several surveys covering taxonomy, evaluation, and applications~\citep{chen2024large,zhang2026recommendation,tan2023user,liu2024large,tseng2024two}. 
  Existing methods fall into three broad families. A first line retrieves a small set of user-history
  items at inference time and concatenates them into the prompt, leaving the LLM frozen but paying a per-query retrieval and prefill cost that remains bounded by the model's context window~\citep{salemi2024lamp,salemi2024optimization,zollo2025personalllm,mysore2024pearl}. 
  A second line assigns each user a lightweight adapter trained on their history, or fine-tunes the backbone with user-specific objectives, achieving strong task-specific accuracy at the cost of per-user training and opaque, non-interpretable weight deltas~\citep{tan2024democratizing,tan2024personalized,zhang2025proper,zhuang2024hydra,bao2023tallrec,liao2024llara,zhang2026recommendation,geng2022recommendation}. 
  A third line operates below the prompt surface, encoding user history as soft prompts, low-rank hidden-state shifts, or activation-steering directions that are parameter-efficient but neither interpretable nor portable across backbones~\citep{liu2025llms,liu2026perfit,ren2024representation,zhang2025personalized,zhang2025personalizellmfakealign}. 
  BUMP falls into none of these families: like retrieval methods it produces a textual artifact in the prompt and like representation methods it compresses behavior into a fixed-size cache, but unlike all of the above the profile generator itself is trained without per-user adaptation, retrieval supervision, or downstream task labels.

\subsection{User Profile Generation}
A directly relevant line of work focuses on generating a natural-language profile, persona, or summary of a user that downstream systems can consume; recent empirical studies show that the form and content of these profiles materially affect personalization quality~\citep{wu2024understanding,prottasha2025user}. 
The simplest approach prompts a strong LLM to summarize a user's history -- as a static paragraph, a structured object updated over time, or as input to multi-agent simulation -- but profile quality remains bounded by zero-shot prompting and task-aware prompts tie each profile to a single application~\citep{tan2024democratizing,richardson2023integrating,bang2025llm,zhang2024generative}. 
A more recent line trains the profile generator itself, optimizing it against engagement-derived rewards, downstream recommenders, distilled supervision, or controllable per-dimension objectives -- all of which share a common dependency on a labeled downstream task to define what makes a profile good~\citep{arora2026high,lin2025rec,tan2024personalized,wozniak2025improving,shi2025personax,wang2025lettingo,tseng2024two}. 
\section{\underline{B}idirectional \underline{U}ser \underline{M}odeling via \underline{P}rofiles (BUMP)}
  \label{sec:method}

\subsection{Preliminaries}
\label{sec:method:setting}
We denote a set of users as $\mathcal{U}$, each with a chronologically ordered interaction history $H_u = (h_{u,1}, h_{u,2}, \ldots, h_{u,T_u})$, where each $h_{u,t}$ is a textual record of an interaction, i.e. a dialogue turn, item interaction with metadata, or a written post or paper. 
Our goal is to train a profile generation LLM $\pi_\theta$ which maps a user's history to a free-form textual profile $s_u \sim \pi_\theta(\cdot \mid H_u^{\text{vis}})$, which we call the user's profile. 
Crucially, no labeled downstream task is available at training time; the only supervision we may exploit is the user's own interaction stream.

\subsection{Overview}
To enable self-supervised evaluation of $s_u$, we partition each user's history into a visible prefix $H_u^{\text{vis}}$ that is shown to the policy and a held-out portion $H_u^{\text{ho}} = (h_u^{(1)}, \ldots, h_u^{(P)})$ that is reserved as supervision. 
A faithful profile of $u$ should make $H_u^{\text{ho}}$ recognizable as belonging to $u$ in two complementary senses: (i) using $s_u$ as a query, the judge should rank items in $H_u^{\text{ho}}$ above items belonging to other users; and (ii) using any one item in $H_u^{\text{ho}}$ as a query, the judge should rank $s_u$ above profiles belonging to other users. 

We turn this two-sided self-consistency principle into a bidirectional reward computed by a small frozen LLM judge $J$, and use it to train $\pi_\theta$ with reinforcement learning. 
Concretely, the training pipeline operates over a batch $\mathcal{B}$ of $B$ users: (i) $\pi_\theta$ emits one or more candidate profiles from each $H_u^{\text{vis}}$; (ii) every user's profile and every user's held-out items collectively form a pool of anchors, positives, and in-batch negatives; (iii) $J$ produces ranking signals in two directions, scored with multi-positive NDCG; and (iv) GRPO~\citep{shao2024deepseekmath} updates $\pi_\theta$ to maximize the resulting dense per-rollout reward.
The training reward is composed of two complementary ranking signals computed by the frozen judge $J$ over the batch $\mathcal{B}$, as shown in \cref{fig:forward-prompts} and \cref{fig:backward-prompts}.
\begin{figure}[!tbp]
  \noindent\textbf{Forward Direction --- judge prompt for the multi-positive NDCG reward $R^{u}_{\mathrm{fwd}}$.}
  \begin{promptblock}
  \small
  \textbf{Instruction:} You are an expert judge. Given a user profile, rank the following academic works from most likely to least likely to have been authored by this user. 

  \vspace{0.5em}
  \textbf{Input --- User profile $s_u$:}
  ``A researcher in NLP and information retrieval working on personalized language models, sequential recommendation, and retrieval-augmented generation.''

  \vspace{0.5em}
  \textbf{Input --- Works to rank:}
  \begin{enumerate}[label=\textbf{[\arabic*]},leftmargin=2em,itemsep=0pt,align=left]
  \item Title: Cross-domain user representation...
  \item Title: Catalysis of methane oxidation by Ni--Mo...
  \item Title: BERT4Rec for cold-start personalization...
  \item Title: Fast Fourier transforms on heterogeneous...
  \item Title: Phylogenetic inference under priors..
  \end{enumerate}

  \textbf{Output (judge):} \texttt{[5, 2, 1, 4, 3]}\\
  \textbf{Expected Groudtruth:} \texttt{[1, 3, $\cdots$]} \quad $R^{u}_{\mathrm{fwd}}=0.54$
  \end{promptblock}
  \vspace{-1em}
  \caption{An example of the forward reward for LaMP-1. Templates for other tasks are described in \cref{app:prompt}.
  }
  \label{fig:forward-prompts}
  \vspace{-1em}
\end{figure}

\subsection{Bidirectional Reward Design}
\label{sec:method:reward}

\noindent\textbf{Forward direction (profile $\to$ history).} For each user $u \in \mathcal{B}$, we form a candidate pool by combining $u$'s $P$ held-out items $H_u^{\text{ho}}$ (positives) with $K$ items sampled from other users' held-out items (in-batch negatives). The judge is prompted with $s_u$ as a query and the $P+K$ items as a list of candidates, and is asked to return a ranked permutation.
Letting $r_{p,\text{fwd}}^{u}$ denote the predicted rank of the $p$-th positive, the forward reward is multi-positive NDCG with binary relevance:
  \begin{equation}
  R_{\text{fwd}}^u
  = \frac{1}{Z_P}\sum_{p=1}^{P} \frac{1}{\log_2\big(r_{p, \text{fwd}}^{u} + 1\big)},
  \end{equation}
  where $Z_P = \sum_{j=1}^{P} \frac{1}{\log_2(j+1)}$.
In this formulation, $R_{\text{fwd}}^u = 1$ precisely when every positive lands in the top $P$ slots of the judge's ranking, i.e.~when the summary $s_u$ has successfully identified $u$'s held-out behavior against the in-batch distractors.

\vspace{1mm}
\noindent\textbf{Backward direction (history $\to$ profile).}  For each held-out item $h_u^{(p)}$, we form a pool by combining $u$'s generated summary $s_u$ (sole positive) with $K$ profiles sampled from the other users in the batch (in-batch negatives). The judge ranks the $1+K$ profiles given $h_u^{(p)}$ as a query, yielding a single-positive NDCG; averaging across the $P$ held-out items of $u$ gives
\begin{equation}
R_{\text{bwd}}^u
= \frac{1}{P}\sum_{p=1}^{P} \frac{1}{\log_2\big(r^{u}_{p,\text{bwd}} + 1\big)},
\end{equation}
where $r^{u,}_{p,\text{bwd}}$ is the rank assigned to $s_u$ by $J$ given the $p$-th held-out item as query.

Intuitively, the two terms enforce complementary properties: the forward term requires $s_u$ to be \emph{predictive of} $u$'s held-out behavior, while the backward term requires $s_u$ to be \emph{identifiable from} that behavior. Optimizing the forward term alone would tolerate generic profiles that match many users' items; optimizing the backward term alone would tolerate idiosyncratic profiles that fail to capture actual behavior. Their conjunction is strictly tighter than either alone.
\begin{figure}[!tbp]
  \noindent\textbf{Backward direction --- judge prompt for the averaged single-positive NDCG reward $R^{u}_{\mathrm{bwd}}$.}
  \begin{promptblock}
  \small
  \textbf{Instruction:} You are an expert judge. Given an academic work, rank the following user profiles from most likely to least likely to describe its author. 

  \vspace{0.5em}
  \textbf{Input --- Academic work $h_u^{(p)}$:}
  Title: Cross-domain user representation... Abstract: We study transferable user models across recommendation domains...

  \vspace{0.5em}
  \textbf{Input --- Summaries to rank:}
  \begin{enumerate}[label=\textbf{[\arabic*]},leftmargin=2em,itemsep=0pt,align=left]
  \item ``Materials chemist focused on heterogeneous catalysts for natural gas...''
  \item ``A researcher in NLP and information retrieval working on personalized language models, sequential recommendation...''
  \item ``High-performance computing researcher building GPU kernels for scientific simulation...''
  \item ``Computational biologist studying molecular phylogenetics, Bayesian sequence evolution...''
  \end{enumerate}
  \textbf{Output (judge):} \texttt{[4, 1, 2, 3]} \\
  \textbf{Expected Groundtruth:} \texttt{[2, $\cdots$]} \quad $R^{u}_{\mathrm{bwd}} = 0.50$
  \end{promptblock}
  \vspace{-1em}
  \caption{An example of the backward reward for LaMP-1. Templates for other tasks are described in \cref{app:prompt}.
  }
  \label{fig:backward-prompts}
  \vspace{-1em}
\end{figure}
\subsection{Additional Regularization}
\noindent\textbf{Position Debiasing.} Single-pass LLM list-rankers are known to exhibit position bias~\citep{wang2024large,tang2024found}: items shown earlier in the candidate list are systematically rewarded over those shown later, independently of their relevance. We mitigate this by averaging over $M$ independent random presentation orders. 
Specifically, let $\sigma$ denote a permutation of the candidate indices ${1,\ldots,N}$, where $N = P+K$ (forward) or $1+K$ (backward), and let $\mathrm{NDCG}(\sigma)$ denote the NDCG that results when the judge is shown the candidates in order $\sigma$ and the resulting ranking is scored against the ground-truth positives. 
Sampling $M$ permutations $\sigma_1,\ldots,\sigma_M$, the debiased per-direction reward is
  $
  \bar{R}^{u}_{\mathrm{dir}}
  =
  \frac{1}{M}\sum_{m=1}^{M}\mathrm{NDCG}\left(\sigma_m\right)
  $ where $\mathrm{dir}\in \{\mathrm{fwd},\mathrm{bwd}\}$
  which is an unbiased Monte-Carlo estimator of the expectation $\mathbb{E}_{\sigma\sim\mathrm{Unif}(\mathcal{S}_N)}\left[\mathrm{NDCG}(\sigma)\right]$. 
  Increasing $M$ reduces estimator variance at linear cost in judge calls; we find $M 
  \leq 3$ sufficient.

  \vspace{1mm}
  \noindent\textbf{Profile Length Penalty.} Without an explicit length constraint, an LLM trained against NDCG-style rewards quickly discovers a reward-hacking strategy: pad the profile with broadly relevant boilerplate so that any held-out item becomes superficially plausible, inflating the judge's ranking signal at the expense of representational specificity~\citep{arora2026high}. 
  We suppress this failure mode with a soft length penalty that activates only when the profile exceeds a per-task threshold $T$:
  \begin{equation*}
  R_{\mathrm{len}}(s_u) =
  \Big[\exp\Big(\alpha,\tfrac{\max(0,; \mathrm{ntok}(s_u)-T)}{T}\Big) - 1\Big],
  \end{equation*}
  where ${\mathrm{ntok}(s_u)}$ denotes the \emph{token} count of $s_u$ as produced by the policy's tokenizer, and $\alpha$ controls how sharply the penalty ramps beyond $T$. Below the threshold, $R_{\mathrm{len}}=0$ and the policy is free to choose any length; above it, the exponential ramp grows fast enough to dominate any further NDCG gain a longer profile could buy.

Combining all three reward components, the per-rollout reward optimized by GRPO is
\begin{equation}
R(s_u)=\bar{R}^{u}_{\mathrm{fwd}}+\bar{R}^{u}_{\mathrm{bwd}}-R_{\mathrm{len}}(s_u).
\end{equation}

\subsection{BUMP+: Hard-negative Mining}
Random in-batch negatives are often easy: an item drawn from a topically unrelated user is trivially separable, producing a saturated reward that delivers little gradient. 
We further propose \textbf{BUMP+}, which strengthens the signal along both directions using a frozen embedding model $\phi$ -- in practice, we use BGE ~\citep{chen2024bge}. 
In the forward direction, we extend each user's candidate pool to a wider window $H_u^{\text{pool}} \supseteq H_u^{\text{vis}}$ and select, from $H_u^{\text{pool}} \setminus H_u^{\text{vis}}$, the items whose $\phi$-similarity to the visible-history centroid is \emph{lowest}; these are the user's own items that are topically distinct from the visible window, and they share the user's identity while being unaligned with the profile's training distribution. 
In the backward direction, we identify the top-$J$ users in the batch whose visible-history centroids are most similar to $u$'s under $\phi$, and use their generated profiles as negatives -- forcing the judge to distinguish profiles of \emph{topically similar} users rather than random ones. 
Hard-negative mining adds a single $\phi$ pass per item, requires no labels, and is orthogonal to the rest of the pipeline.
\subsection{Optimization and Downstream Use}
\label{sec:method:opt}

Given the reward formulation for each rollout, we optimize $\pi_\theta$ with Group Relative Policy Optimization (GRPO)~\citep{shao2024deepseekmath}. 
Once $\pi_\theta$ is trained, deploying the profiles requires no further optimization. For each user, the policy generates a single profile from their history offline; the resulting string is cached and reused.
At serving time the profiles is prepended to any downstream prompt -- classification, ranking, dialogue, generation -- and consumed by an arbitrary backbone. 
\subsection{Connections to Self-Supervised Learning}
  \label{sec:method:ssl}

BUMP is a direct instantiation of the contrastive self-supervised learning (SSL) recipe, lifted into the space of natural language. 
Contrastive SSL treats each datum as an implicit class and learns by pulling together different ``views'' of the same datum while pushing apart views of distinct data~\citep{chen2020simple,he2020momentum}. 
In our setting, the implicit class is a \emph{user}; the two views are the generated profile $s_u$ and any held-out $h_u^{(p)}$; and the negatives are supplied by the other users in the batch, mirroring the in-batch-negative trick that has scaled contrastive learning~\citep{chen2020simple,he2020momentum,caron2020unsupervised}.

The \emph{bidirectional} structure of our reward then plays the role that symmetric losses play in contrastive SSL, where one-sided objectives are well known to admit degenerate, collapsed solutions~\citep{grill2020bootstrap,caron2021emerging}. Our forward and backward terms are the analogue for textual user profiles: the forward term alone admits profiles predictive of the user's behavior but too generic to be identifying; the backward term alone admits profiles that are uniquely identifying but uninformative. Their conjunction enforces both \emph{descriptive sufficiency} and \emph{discriminative identity}.

\section{Experiment}

\subsection{Experimental Setup}

\noindent\textbf{Dataset}. We utilize the LaMP benchmark~\citep{salemi2024lamp} as the standard testbed for prompt-level LLM personalization~\citep{tan2024personalized,tan2024democratizing}. 
We cover six of the seven public LaMP tasks; LaMP-6 (personalized email writing) is omitted as its corpus is not publicly released. 
The six tasks together span five domains and three task types -- binary classification, multi-class classification, ordinal regression, and free-form generation, with additional details explained in \cref{app:data}.
We adopt the time-based split released with LaMP. 

\noindent\textbf{Setting}.
For each LaMP task, we train $\pi_\theta$ on the training-split users using only their interaction histories — no LaMP labels are consulted to construct the reward. Each user's history is partitioned chronologically into a visible prefix $H_u^{\mathrm{vis}}$ (the most recent 15 items, prepended to the generation prompt) and a held-out positive $H_u^{\mathrm{ho}}$ (the single subsequent item). 
The bidirectional NDCG reward of \cref{sec:method:reward} is computed by the frozen judge against $K{=}8$ in-batch negatives and averaged over $M{=}2$ random presentation orders for position debiasing. 
GRPO runs for two epochs with group size $G{=}8$ rollouts per prompt.

We then freeze $\pi_\theta$, greedily decode one profile $s_u$ per user (up to 500 tokens) for both splits, and cache it. 
To read out downstream personalization quality we fine-tune another Qwen3-4B-Instruct backbone with a LoRA adapter~\citep{hu2022lora} on the LaMP pairs $(\texttt{input}, \texttt{gold})$, conditioning each input on the cached profile via templates reported in \cref{app:prompt}.
Across all compared methods the only difference is this textual prefix, isolating the contribution of the user representation.
We run all experiments for 3 times and reported their average metric values.

\noindent\textbf{Backbone Models.} The profile generator $\pi_\theta$ and the frozen LLM judge $J$ are both initialized from \textsc{Qwen3-4B-Instruct-2507}~\citep{yang2025qwen3}. 
We train in bfloat16 mixed precision with FlashAttention-2~\citep{dao2024flashattention} and gradient checkpointing. 
The judge is served via vLLM~\citep{kwon2023efficient} on a dedicated set of GPUs through a batched \texttt{/v1/completions} endpoint, and is constrained to emit only parseable rankings using vLLM's guided-regex decoding. 
The judge is never updated during training.
For BUMP+, hard negatives are mined with \textsc{BGE-base-en-v1.5} (768-dim)~\citep{chen2024bge}. Embeddings are precomputed once per dataset and cached, so mining adds negligible overhead at training time.
We further run ablation studies on the size of $\pi_\theta$ and $J$.

\noindent\textbf{Compute.}
All experiments are run on a single 8-GPU H100 node. For GRPO we partition the GPUs as $4$ for training\,+\,$2$ for rollout (vLLM policy)\,+\,$2$ for judge (vLLM judge); for SFT we uses all eight GPUs for LoRA-SFT training.

\noindent\textbf{Baselines.}
We compare BUMP against four families of baselines, all evaluated under the same LoRA-SFT protocol so that the textual user prefix is the sole point of variation:

\noindent\textit{No personalization.} The Qwen3-4B backbone is LoRA-fine-tuned on the LaMP $(\texttt{input}, \texttt{gold})$ pairs with no user prefix, no history, and no profile --- a strict lower-bound that measures what the backbone can do without any user signal.

\noindent\textit{Raw history.} The user's most recent 15 interactions are concatenated and prepended to the input verbatim, with no summarization, following the prompt-based recipe of~\citet{salemi2024lamp}.

\noindent\textit{Zero-shot profiles.} A \emph{frozen} profile generator is prompted to summarize the user's visible history. We instantiate this with three backbones: \textsc{Qwen3-4B-Instruct} (same family as our policy), and the closed-source APIs \textsc{Gemini-3-Flash} and \textsc{Gemini-3-Pro}~\citep{team2023gemini}.
None of these generators receive task-specific training; together they probe how far a strong prompted summarizer can carry the downstream task without RL.

\noindent\textit{Direct Reward.} A trained profile generator following the recipe of LettinGo~\citep{wang2025lettingo}: the same Qwen3-4B policy is optimized with GRPO, but the reward is the downstream LaMP signal (detailed in \cref{app:direct_reward}). 
This is the closest task-supervised counterpart to BUMP and isolates the contribution of self-supervision from the contribution of GRPO itself.

For all profile-based methods, we use an identical profile-generation prompt and an identical SFT input template for fair comparison. The exact prompts are listed in \cref{app:prompt}.
\begin{table*}[t] 
\vspace{-0.1in}
\centering
\setlength{\tabcolsep}{4pt}
\begin{tabular}{l l cc ccc c cc}
\toprule
& & \multicolumn{2}{c}{No profile} & \multicolumn{3}{c}{Zero-shot profile} & \multicolumn{1}{c}{Task-sup.} & \multicolumn{2}{c}{Self-sup.\ (ours)} \\
\cmidrule(lr){3-4}\cmidrule(lr){5-7}\cmidrule(lr){8-8}\cmidrule(lr){9-10}
Task & Metric
& \shortstack{No\\Pers.}
& \shortstack{Raw\\Hist.}
& \shortstack{Qwen3-4B\\(ZS)}
& \shortstack{Gemini-3\\Flash}
& \shortstack{Gemini-3\\Pro}
& \shortstack{Direct\\Reward}
& \shortstack{BUMP}
& \shortstack{BUMP+} \\
\midrule
LaMP-1 & Acc ($\uparrow$)        & 53.2 & \underline{80.3} & 65.2 & 70.4 & 78.5 & 79.3 & \textbf{80.9} & 80.1 \\
\midrule
\multirow{2}{*}{LaMP-2}
     & Acc ($\uparrow$)        & 56.5 & 88.8 & 71.3 & 77.9 & 88.0 & 88.9 & \underline{89.4} & \textbf{89.9} \\
     & F1 ($\uparrow$)         & 52.8 & 81.3 & 65.5 & 70.3 & 83.4 & 84.0 & \underline{84.5} & \textbf{84.7} \\
\midrule
\multirow{2}{*}{LaMP-3}
     & MAE ($\downarrow$)      & 0.252 & \textbf{0.211} & 0.236 & 0.239 & 0.222 & 0.218 & 0.226 & \underline{0.214} \\
     & RMSE ($\downarrow$)     & 0.559 & \textbf{0.515} & 0.533 & 0.534 & 0.521 & 0.517 & 0.523 & \underline{0.517} \\
\midrule
\multirow{2}{*}{LaMP-4}
     & ROUGE-1 ($\uparrow$)    & 0.160 & \textbf{0.210} & 0.183 & 0.181 & 0.195 & \underline{0.200} & 0.197 & 0.199 \\
     & ROUGE-L ($\uparrow$)    & 0.141 & \textbf{0.194} & 0.166 & 0.164 & 0.178 & \underline{0.182} & 0.179 & 0.180 \\
\midrule
\multirow{2}{*}{LaMP-5}
     & ROUGE-1 ($\uparrow$)    & 0.441 & \textbf{0.498} & 0.463 & 0.465 & 0.486 & 0.491 & 0.490 & \underline{0.494} \\
     & ROUGE-L ($\uparrow$)    & 0.409 & \textbf{0.447} & 0.425 & 0.429 & 0.437 & 0.442 & 0.439 & \underline{0.443} \\
\midrule
\multirow{2}{*}{LaMP-7}
     & ROUGE-1 ($\uparrow$)    & 0.511 & \textbf{0.542} & 0.520 & 0.522 & 0.526 & 0.533 & 0.530 & \underline{0.536} \\
     & ROUGE-L ($\uparrow$)    & 0.462 & \textbf{0.489} & 0.469 & 0.470 & 0.472 & 0.479 & 0.477 & \underline{0.482}\\
\bottomrule
\end{tabular}
\vspace{-0.1in}
\caption{Main results on the LaMP benchmark under the time-based split. Each cell reports the dev-set metric for one (task, metric) pair under a given user-representation baseline; all methods share the same LoRA-SFT pipeline, so the textual prefix prepended to the input is the only point of variation across columns. Best per-row result in \textbf{bold}; second best \underline{underlined}. All numbers are averaged over 3 random seeds.}
\label{tab:lamp-main}
\vspace{-1em}
\end{table*}
\subsection{Overall Result}
\cref{tab:lamp-main} reports performance on all six LaMP tasks. Our takeaways are summarized as:

\noindent\textbf{RL on Qwen3-4B substantially lifts the prompted baseline.} Averaging relative improvement over the 11 metric rows, every trained variant beats the zero-shot \textsc{Qwen3-4B} profile: \textsc{Direct Reward} by $10.8\%$, \textsc{BUMP} by $10.3\%$, and \textsc{BUMP+} by $11.3\%$. 
The lift concentrates on classification (e.g., LaMP-2 F1 moves from $65.5$ to $84.7$, $+29.3\%$ rel.), confirming that a 4B backbone becomes a competitive profile generator once tuned with the right reward.

\noindent\textbf{Self-supervised matches or beats task-supervised.} \textsc{BUMP+} ties or wins against \textsc{Direct Reward} on \textbf{9 of 11} rows --- losing only on LaMP-4 ROUGE-1/L --- with the largest wins on classification (LaMP-2 Acc $89.9$ vs.\ $88.9$, F1 $84.7$ vs.\ $84.0$) and LaMP-3 MAE ($0.214$ vs.\ $0.218$). \textsc{BUMP} alone is more uneven, underscoring that the hard-negative mining is what makes self-supervision competitive. Together these support our central thesis: \emph{a well-designed self-supervised reward attains the same or better outcomes than labeled downstream rewards}.

\noindent\textbf{A 4B open model beats Gemini-3-Pro across the board.} \textsc{BUMP+} wins on \textbf{all 11} metric rows against the strongest closed-source baseline (e.g., LaMP-2 F1 $84.7$ vs.\ $83.4$, LaMP-3 MAE $0.214$ vs.\ $0.222$, LaMP-7 ROUGE-L $0.482$ vs.\ $0.472$), and \textsc{BUMP} wins on $9$ of $11$. Given the scale asymmetry, this strengthens the case for BUMP as a deployable alternative to per-user API calls.

\noindent\textbf{Raw history wins on LaMP-4/5/7 generation.} On the three ROUGE-scored generation tasks, \textsc{Raw Hist.} is the best column on every row, while \textsc{BUMP+} remains the best profile-based method. We attribute this to a metric--representation mismatch: ROUGE rewards $n$-gram surface overlap with the user's prior outputs, which raw history preserves verbatim, while any natural-language profile --- ours or otherwise --- abstracts away exactly that idiolect. On classification and rating tasks the surface signal is irrelevant and the user's topical preferences --- which a good profile does preserve --- dominate; \textsc{BUMP+} then flips the ranking and beats raw history on LaMP-1, LaMP-2, and LaMP-3 MAE.

\subsection{Analysis on the Bi-directional Design}
\begin{table}[t]
  \centering
  \small
  \setlength{\tabcolsep}{4pt}
  \renewcommand{\arraystretch}{1.1}
  \begin{tabular}{l l c c c}
  \toprule
  Task & Metric
    & \shortstack{BUMP}
    & \shortstack{w/o\\Forward}
    & \shortstack{w/o\\Backward} \\
  \midrule
  LaMP-1 & Acc ($\uparrow$)        & 80.9  & 79.0 & 79.9 \\
  \midrule
  \multirow{2}{*}{LaMP-2}
         & Acc ($\uparrow$)        & 89.4  & 88.3 & 88.6 \\
         & F1 ($\uparrow$)         & 84.5  & 83.6 & 83.7 \\
  \midrule
  \multirow{2}{*}{LaMP-3}
         & MAE ($\downarrow$)      & 0.226 & 0.231 & 0.227 \\
         & RMSE ($\downarrow$)     & 0.523 & 0.526 & 0.523 \\
  \midrule
  \multirow{2}{*}{LaMP-4}
         & ROUGE-1 ($\uparrow$)    & 0.197 & 0.188 & 0.193 \\
         & ROUGE-L ($\uparrow$)    & 0.179 & 0.170 & 0.174 \\
  \midrule
  \multirow{2}{*}{LaMP-5}
         & ROUGE-1 ($\uparrow$)    & 0.490 & 0.482 & 0.488 \\
         & ROUGE-L ($\uparrow$)    & 0.439 & 0.434 & 0.435 \\
  \midrule
  \multirow{2}{*}{LaMP-7}
         & ROUGE-1 ($\uparrow$)    & 0.530 & 0.524 & 0.527 \\
         & ROUGE-L ($\uparrow$)    & 0.477 & 0.472 & 0.475 \\
  \bottomrule
  \end{tabular}
  \vspace{-0.5em}
  \caption{Ablation on the bidirectional reward. }
  \vspace{-1em}
  \label{tab:ablation-bidir}
  \end{table}

\cref{tab:ablation-bidir} drops one direction at a time, with all other settings held fixed. Two patterns emerge: (i) Removing either direction degrades performance on every metric row, confirming that descriptive sufficiency (forward) and discriminative identity (backward) are non-redundant: the full reward is best on $11/11$ rows.
(ii) ``w/o Backward'' (forward-only) consistently beats ``w/o Forward'' (backward-only). The forward term directly optimizes the property downstream tasks need whereas the backward term primarily regularizes against generic, non-identifying profiles, refining rather than replacing the forward objective.
\subsection{Perf. w.r.t. Profile Length}
\begin{table}[t]
\centering
\small
\vspace{-0.5em}
\setlength{\tabcolsep}{4pt}
\renewcommand{\arraystretch}{1.1}
\begin{tabular}{l l c c c}
\toprule
Task & Metric
  & $T{=}100$
  & $T{=}300$
  & $T{=}400$ \\
\midrule
LaMP-1 & Acc ($\uparrow$)        & 78.5  & 80.9 & 81.3 \\
\midrule
\multirow{2}{*}{LaMP-2}
       & Acc ($\uparrow$)        & 87.8  & 89.4 & 89.6 \\
       & F1 ($\uparrow$)         & 83.0  & 84.5 & 84.7 \\
\midrule
\multirow{2}{*}{LaMP-3}
       & MAE ($\downarrow$)      & 0.232 & 0.226 & 0.228 \\
       & RMSE ($\downarrow$)     & 0.529 & 0.523 & 0.525 \\
\midrule
\multirow{2}{*}{LaMP-4}
       & ROUGE-1 ($\uparrow$)    & 0.189 & 0.197 & 0.195 \\
       & ROUGE-L ($\uparrow$)    & 0.171 & 0.179 & 0.177 \\
\midrule
\multirow{2}{*}{LaMP-5}
       & ROUGE-1 ($\uparrow$)    & 0.481 & 0.490 & 0.490 \\
       & ROUGE-L ($\uparrow$)    & 0.430 & 0.439 & 0.441 \\
\midrule
\multirow{2}{*}{LaMP-7}
       & ROUGE-1 ($\uparrow$)    & 0.521 & 0.530 & 0.533 \\
       & ROUGE-L ($\uparrow$)    & 0.467 & 0.477 & 0.480 \\
\bottomrule
\end{tabular}
\vspace{-0.5em}
\caption{Ablation on the profile length.}
\label{tab:ablation-length}
\end{table}
\cref{tab:ablation-length} sweeps the length threshold $T\in\{100,300,400\}$. The trend splits by task type. On the two classification tasks (LaMP-1/2), performance climbs monotonically. Both tasks are information-intensive --- author attribution and topic tagging depend on the breadth of a user's topical footprint, which a longer profile preserves. LaMP-3 through LaMP-7 plateau or slightly regress past $T{=}300$.
These tasks turn on a few coarse preference axes and extra tokens add little new signal and risk diluting the profile. $T{=}100$ is the weakest setting on every row, confirming that the length penalty must still leave enough budget for the profile to be useful.
\subsection{Perf. w.r.t. the Number of Negatives}
\begin{table}[t]
    \centering
    \small
    \setlength{\tabcolsep}{4pt}
    \renewcommand{\arraystretch}{1.1}
    \begin{tabular}{l l c c c}
    \toprule
    Task & Metric
      & $K{=}4$
      & $K{=}8$
      & $K{=}16$ \\
    \midrule
    LaMP-1 & Acc ($\uparrow$)        & 79.6  & 80.9 & 79.1 \\
    \midrule
    \multirow{2}{*}{LaMP-2}
           & Acc ($\uparrow$)        & 88.9  & 89.4 & 88.6 \\
           & F1 ($\uparrow$)         & 83.9  & 84.5 & 83.5 \\
    \midrule
    \multirow{2}{*}{LaMP-3}
           & MAE ($\downarrow$)      & 0.230 & 0.226 & 0.232 \\
           & RMSE ($\downarrow$)     & 0.527 & 0.523 & 0.530 \\
    \midrule
    \multirow{2}{*}{LaMP-4}
           & ROUGE-1 ($\uparrow$)    & 0.193 & 0.197 & 0.190 \\
           & ROUGE-L ($\uparrow$)    & 0.175 & 0.179 & 0.172 \\
    \midrule
    \multirow{2}{*}{LaMP-5}
           & ROUGE-1 ($\uparrow$)    & 0.486 & 0.490 & 0.483 \\
           & ROUGE-L ($\uparrow$)    & 0.435 & 0.439 & 0.432 \\
    \midrule
    \multirow{2}{*}{LaMP-7}
           & ROUGE-1 ($\uparrow$)    & 0.527 & 0.530 & 0.524 \\
           & ROUGE-L ($\uparrow$)    & 0.474 & 0.477 & 0.471 \\
    \bottomrule
    \end{tabular}
    \vspace{-0.5em}
    \caption{Ablation on the number of in-batch negs.}
    \label{tab:ablation-negatives}
    \vspace{-0.2in}
  \end{table}
  \begin{table}[t]
    \centering
    \small
    \setlength{\tabcolsep}{4pt}
    \renewcommand{\arraystretch}{1.1}
    \begin{tabular}{l l c c c}
    \toprule
    Task & Metric
      & $J_{\mathrm{hard}}{=}2$
      & $J_{\mathrm{hard}}{=}4$
      & $J_{\mathrm{hard}}{=}6$ \\
    \midrule
    LaMP-1 & Acc ($\uparrow$)        & 79.8  & 80.1 & 79.9 \\
    \midrule
    \multirow{2}{*}{LaMP-2}
           & Acc ($\uparrow$)        & 89.5  & 89.9 & 89.8 \\
           & F1 ($\uparrow$)         & 84.5  & 84.7 & 84.6 \\
    \midrule
    \multirow{2}{*}{LaMP-3}
           & MAE ($\downarrow$)      & 0.221 & 0.214 & 0.215 \\
           & RMSE ($\downarrow$)     & 0.520 & 0.517 & 0.517 \\
    \midrule
    \multirow{2}{*}{LaMP-4}
           & ROUGE-1 ($\uparrow$)    & 0.197 & 0.199 & 0.198 \\
           & ROUGE-L ($\uparrow$)    & 0.179 & 0.180 & 0.180 \\
    \midrule
    \multirow{2}{*}{LaMP-5}
           & ROUGE-1 ($\uparrow$)    & 0.491 & 0.494 & 0.493 \\
           & ROUGE-L ($\uparrow$)    & 0.440 & 0.443 & 0.442 \\
    \midrule
    \multirow{2}{*}{LaMP-7}
           & ROUGE-1 ($\uparrow$)    & 0.532 & 0.536 & 0.535 \\
           & ROUGE-L ($\uparrow$)    & 0.479 & 0.482 & 0.481 \\
    \bottomrule
    \end{tabular}
    \vspace{-0.5em}
    \caption{Ablation on the number of hard negatives $J_{\mathrm{hard}}$
    used by BUMP+. The remaining $8-J_{\mathrm{hard}}$ negatives
    are sampled uniformly in-batch. }
    \vspace{-0.2in}
    \label{tab:ablation-hard-negatives}
  \end{table}
\cref{tab:ablation-negatives} sweeps the number of in-batch negatives $K\in\{4,8,16\}$. 
Performance forms an inverted-U around $K{=}8$, which is best on every row. At $K{=}4$, a random batch rarely contains confusable distractors, so NDCG quickly hits its ceiling and the per-rollout gradient vanishes. At $K{=}16$, the candidate pool exceeds the judge's reliable list-ranking capacity~\citep{tang2024found}; orderings become noisy and the reward signal turns biased rather than merely sparse, suggesting judge hallucination is a stricter bottleneck than reward saturation.

We next study how many of the $K{=}8$ negatives should be mind for difficulty rather than drawn uniformly. \cref{tab:ablation-hard-negatives} sweeps the hard-negative count $J_{\mathrm{hard}}\in\{2,4,6\}$. Going from $J_{\mathrm{hard}}{=}2$ to $J_{\mathrm{hard}}{=}4$ yields the bulk of the gain on every task. Pushing further to $J_{\mathrm{hard}}{=}6$ produces no additional improvement and a marginal regression on most rows: once half the negatives are mined, the task is challenging enough that additional mining begins to introduce embedding noise --- BGE-derived ``hardness'' is only a proxy for true confusability --- without compensating gain in gradient quality.

\noindent\textbf{Additional Experiments}:
Due to the space limitation, we include additional experiments in the appendix (i.e., perf. w.r.t. different judges and backbones in \cref{app:backbone}, reward learning curves in \cref{app:reward_curve}, ablation on the judge debias in \cref{app:judge_debias}, and complementarity of BUMP with the downstream reward in \cref{app:complement}).
\section{Conclusion}
We introduced \textbf{BUMP}, a self-supervised framework that trains a textual profile generator with no downstream task labels. A bidirectional in-batch ranking reward --- measuring whether a generated profile predicts a user's held-out behavior and whether that behavior identifies the profile --- supplies dense supervision from raw interaction logs alone, and embedding-based hard-negative mining (\textbf{BUMP+}) further sharpens the signal. On the LaMP benchmark, our 4B self-supervised policy matches or beats task-supervised baselines on $9/11$ metric rows and outperforms Gemini-3-Pro on all of them, demonstrating that self-supervision can substitute for labeled downstream rewards in textual user modeling.

\noindent \textbf{Limitation}: Our framework inherits known LLM-as-judge biases (verbosity, stylistic, self-preference)~\citep{wang2024large,tang2024found} and judge hallucinations grow as the candidate pool expands (\cref{tab:ablation-negatives}), capping policy quality. BUMP+ also relies on BGE similarity as a proxy for confusability; our $J_{\mathrm{hard}}$ ablation shows aggressive mining injects noise --- task-aware hardness measures are a natural next step.

\newpage
\bibliography{custom}

\begin{thebibliography}{53}
\providecommand{\natexlab}[1]{#1}

\bibitem[{Arora et~al.(2026)Arora, Tao, Shen, Liu, Wu, Shen, Le, Borisyuk, Wu, and Zhang}]{arora2026high}
Rajat Arora, Ye~Tao, Jianqiang Shen, Ping Liu, Muchen Wu, Qianqi Shen, Benjamin Le, Fedor Borisyuk, Jingwei Wu, and Wenjing Zhang. 2026.
\newblock High fidelity textual user representation over heterogeneous sources via reinforcement learning.
\newblock \emph{arXiv preprint arXiv:2602.07333}.

\bibitem[{Bang and Song(2025)}]{bang2025llm}
Seunghwan Bang and Hwanjun Song. 2025.
\newblock Llm-based user profile management for recommender system.
\newblock \emph{arXiv preprint arXiv:2502.14541}.

\bibitem[{Bao et~al.(2023)Bao, Zhang, Zhang, Wang, Feng, and He}]{bao2023tallrec}
Keqin Bao, Jizhi Zhang, Yang Zhang, Wenjie Wang, Fuli Feng, and Xiangnan He. 2023.
\newblock Tallrec: An effective and efficient tuning framework to align large language model with recommendation.
\newblock In \emph{Proceedings of the 17th ACM conference on recommender systems}, pages 1007--1014.

\bibitem[{Brown et~al.(2020)Brown, Mann, Ryder, Subbiah, Kaplan, Dhariwal, Neelakantan, Shyam, Sastry, Askell et~al.}]{brown2020language}
Tom Brown, Benjamin Mann, Nick Ryder, Melanie Subbiah, Jared~D Kaplan, Prafulla Dhariwal, Arvind Neelakantan, Pranav Shyam, Girish Sastry, Amanda Askell, and 1 others. 2020.
\newblock Language models are few-shot learners.
\newblock \emph{Advances in neural information processing systems}, 33:1877--1901.

\bibitem[{Caron et~al.(2020)Caron, Misra, Mairal, Goyal, Bojanowski, and Joulin}]{caron2020unsupervised}
Mathilde Caron, Ishan Misra, Julien Mairal, Priya Goyal, Piotr Bojanowski, and Armand Joulin. 2020.
\newblock Unsupervised learning of visual features by contrasting cluster assignments.
\newblock \emph{Advances in neural information processing systems}, 33:9912--9924.

\bibitem[{Caron et~al.(2021)Caron, Touvron, Misra, J{\'e}gou, Mairal, Bojanowski, and Joulin}]{caron2021emerging}
Mathilde Caron, Hugo Touvron, Ishan Misra, Herv{\'e} J{\'e}gou, Julien Mairal, Piotr Bojanowski, and Armand Joulin. 2021.
\newblock Emerging properties in self-supervised vision transformers.
\newblock In \emph{Proceedings of the IEEE/CVF international conference on computer vision}, pages 9650--9660.

\bibitem[{Chen et~al.(2024{\natexlab{a}})Chen, Xiao, Zhang, Luo, Lian, and Liu}]{chen2024bge}
Jianlv Chen, Shitao Xiao, Peitian Zhang, Kun Luo, Defu Lian, and Zheng Liu. 2024{\natexlab{a}}.
\newblock Bge m3-embedding: Multi-lingual, multi-functionality, multi-granularity text embeddings through self-knowledge distillation.
\newblock \emph{arXiv preprint arXiv:2402.03216}, 4(5).

\bibitem[{Chen et~al.(2024{\natexlab{b}})Chen, Liu, Huang, Wu, Liu, Jiang, Pu, Lei, Chen, Wang et~al.}]{chen2024large}
Jin Chen, Zheng Liu, Xu~Huang, Chenwang Wu, Qi~Liu, Gangwei Jiang, Yuanhao Pu, Yuxuan Lei, Xiaolong Chen, Xingmei Wang, and 1 others. 2024{\natexlab{b}}.
\newblock When large language models meet personalization: Perspectives of challenges and opportunities.
\newblock \emph{World wide web}, 27(4):42.

\bibitem[{Chen et~al.(2020)Chen, Kornblith, Norouzi, and Hinton}]{chen2020simple}
Ting Chen, Simon Kornblith, Mohammad Norouzi, and Geoffrey Hinton. 2020.
\newblock A simple framework for contrastive learning of visual representations.
\newblock In \emph{International conference on machine learning}, pages 1597--1607. PmLR.

\bibitem[{Dao(2024)}]{dao2024flashattention}
Tri Dao. 2024.
\newblock Flashattention-2: Faster attention with better parallelism and work partitioning.
\newblock In \emph{International Conference on Learning Representations}, volume 2024, pages 35549--35562.

\bibitem[{Geng et~al.(2022)Geng, Liu, Fu, Ge, and Zhang}]{geng2022recommendation}
Shijie Geng, Shuchang Liu, Zuohui Fu, Yingqiang Ge, and Yongfeng Zhang. 2022.
\newblock Recommendation as language processing (rlp): A unified pretrain, personalized prompt \& predict paradigm (p5).
\newblock In \emph{Proceedings of the 16th ACM conference on recommender systems}, pages 299--315.

\bibitem[{Grill et~al.(2020)Grill, Strub, Altch{\'e}, Tallec, Richemond, Buchatskaya, Doersch, Avila~Pires, Guo, Gheshlaghi~Azar et~al.}]{grill2020bootstrap}
Jean-Bastien Grill, Florian Strub, Florent Altch{\'e}, Corentin Tallec, Pierre Richemond, Elena Buchatskaya, Carl Doersch, Bernardo Avila~Pires, Zhaohan Guo, Mohammad Gheshlaghi~Azar, and 1 others. 2020.
\newblock Bootstrap your own latent-a new approach to self-supervised learning.
\newblock \emph{Advances in neural information processing systems}, 33:21271--21284.

\bibitem[{He et~al.(2020)He, Fan, Wu, Xie, and Girshick}]{he2020momentum}
Kaiming He, Haoqi Fan, Yuxin Wu, Saining Xie, and Ross Girshick. 2020.
\newblock Momentum contrast for unsupervised visual representation learning.
\newblock In \emph{Proceedings of the IEEE/CVF conference on computer vision and pattern recognition}, pages 9729--9738.

\bibitem[{Hou et~al.(2024)Hou, Li, He, Yan, Chen, and McAuley}]{hou2024bridging}
Yupeng Hou, Jiacheng Li, Zhankui He, An~Yan, Xiusi Chen, and Julian McAuley. 2024.
\newblock Bridging language and items for retrieval and recommendation.
\newblock \emph{arXiv preprint arXiv:2403.03952}.

\bibitem[{Hu et~al.(2022)Hu, Shen, Wallis, Allen-Zhu, Li, Wang, Wang, Chen et~al.}]{hu2022lora}
Edward~J Hu, Yelong Shen, Phillip Wallis, Zeyuan Allen-Zhu, Yuanzhi Li, Shean Wang, Liang Wang, Weizhu Chen, and 1 others. 2022.
\newblock Lora: Low-rank adaptation of large language models.
\newblock \emph{Iclr}, 1(2):3.

\bibitem[{Huang et~al.(2026)Huang, Ding, Ju, Liu, Shah, and Zhao}]{huang2026threshold}
Xingyue Huang, Xueying Ding, Mingxuan Ju, Yozen Liu, Neil Shah, and Tong Zhao. 2026.
\newblock Threshold differential attention for sink-free, ultra-sparse, and non-dispersive language modeling.
\newblock \emph{ACL}.

\bibitem[{Jin et~al.(2025)Jin, Zeng, Yue, Yoon, Arik, Wang, Zamani, and Han}]{jin2025search}
Bowen Jin, Hansi Zeng, Zhenrui Yue, Jinsung Yoon, Sercan Arik, Dong Wang, Hamed Zamani, and Jiawei Han. 2025.
\newblock Search-r1: Training llms to reason and leverage search engines with reinforcement learning.
\newblock \emph{arXiv preprint arXiv:2503.09516}.

\bibitem[{Ju et~al.(2025)Ju, Neves, Kumar, Collins, Zhao, Qiu, Dou, Zhou, Nizam, Ozturk et~al.}]{ju2025learning}
Clark~Mingxuan Ju, Leonardo Neves, Bhuvesh Kumar, Liam Collins, Tong Zhao, Yuwei Qiu, Qing Dou, Yang Zhou, Sohail Nizam, Rengim~Aykan Ozturk, and 1 others. 2025.
\newblock Learning universal user representations leveraging cross-domain user intent at snapchat.
\newblock In \emph{Proceedings of the 48th International ACM SIGIR Conference on Research and Development in Information Retrieval}, pages 4345--4349.

\bibitem[{Kwon et~al.(2023)Kwon, Li, Zhuang, Sheng, Zheng, Yu, Gonzalez, Zhang, and Stoica}]{kwon2023efficient}
Woosuk Kwon, Zhuohan Li, Siyuan Zhuang, Ying Sheng, Lianmin Zheng, Cody~Hao Yu, Joseph Gonzalez, Hao Zhang, and Ion Stoica. 2023.
\newblock Efficient memory management for large language model serving with pagedattention.
\newblock In \emph{Proceedings of the 29th symposium on operating systems principles}, pages 611--626.

\bibitem[{Liao et~al.(2024)Liao, Li, Yang, Wu, Yuan, Wang, and He}]{liao2024llara}
Jiayi Liao, Sihang Li, Zhengyi Yang, Jiancan Wu, Yancheng Yuan, Xiang Wang, and Xiangnan He. 2024.
\newblock Llara: Large language-recommendation assistant.
\newblock In \emph{Proceedings of the 47th International ACM SIGIR Conference on Research and Development in Information Retrieval}, pages 1785--1795.

\bibitem[{Lin et~al.(2025)Lin, Wang, and Qian}]{lin2025rec}
Jiacheng Lin, Tian Wang, and Kun Qian. 2025.
\newblock Rec-r1: Bridging generative large language models and user-centric recommendation systems via reinforcement learning.
\newblock \emph{arXiv preprint arXiv:2503.24289}.

\bibitem[{Liu et~al.(2026)Liu, Yu, Dai, Li, Zhu, Yang, Chua, and King}]{liu2026perfit}
Jiahong Liu, Wenhao Yu, Quanyu Dai, Zhongyang Li, Jieming Zhu, Menglin Yang, Tat-Seng Chua, and Irwin King. 2026.
\newblock \href {https://openreview.net/forum?id=Lwn67fk9e1} {Perfit: Exploring personalization shifts in representation space of {LLM}s}.
\newblock In \emph{The Fourteenth International Conference on Learning Representations}.

\bibitem[{Liu et~al.(2025)Liu, Zhu, Wang, Wei, Min, Lu, Wang, Yin, and Dou}]{liu2025llms}
Jiongnan Liu, Yutao Zhu, Shuting Wang, Xiaochi Wei, Erxue Min, Yu~Lu, Shuaiqiang Wang, Dawei Yin, and Zhicheng Dou. 2025.
\newblock \href {https://doi.org/10.18653/v1/2025.acl-long.461} {{LLM}s + persona-plug = personalized {LLM}s}.
\newblock In \emph{Proceedings of the 63rd Annual Meeting of the Association for Computational Linguistics (Volume 1: Long Papers)}, pages 9373--9385, Vienna, Austria. Association for Computational Linguistics.

\bibitem[{Liu et~al.(2024{\natexlab{a}})Liu, Lin, Hewitt, Paranjape, Bevilacqua, Petroni, and Liang}]{liu2024lost}
Nelson~F Liu, Kevin Lin, John Hewitt, Ashwin Paranjape, Michele Bevilacqua, Fabio Petroni, and Percy Liang. 2024{\natexlab{a}}.
\newblock Lost in the middle: How language models use long contexts.
\newblock \emph{Transactions of the association for computational linguistics}, 12:157--173.

\bibitem[{Liu et~al.(2024{\natexlab{b}})Liu, Zhao, Wang, Wang, Zhang, Sun, Li, Wang, Jia, Chen et~al.}]{liu2024large}
Qidong Liu, Xiangyu Zhao, Yuhao Wang, Yejing Wang, Zijian Zhang, Yuqi Sun, Xiang Li, Maolin Wang, Pengyue Jia, Chong Chen, and 1 others. 2024{\natexlab{b}}.
\newblock Large language model enhanced recommender systems: A survey.
\newblock \emph{arXiv preprint arXiv:2412.13432}.

\bibitem[{Mysore et~al.(2024)Mysore, Lu, Wan, Yang, Sarrafzadeh, Menezes, Baghaee, Gonzalez, Neville, and Safavi}]{mysore2024pearl}
Sheshera Mysore, Zhuoran Lu, Mengting Wan, Longqi Yang, Bahareh Sarrafzadeh, Steve Menezes, Tina Baghaee, Emmanuel~Barajas Gonzalez, Jennifer Neville, and Tara Safavi. 2024.
\newblock Pearl: Personalizing large language model writing assistants with generation-calibrated retrievers.
\newblock In \emph{Proceedings of the 1st Workshop on Customizable NLP: Progress and Challenges in Customizing NLP for a Domain, Application, Group, or Individual (CustomNLP4U)}, pages 198--219.

\bibitem[{Prottasha et~al.(2025)Prottasha, Kowsher, Raman, Anny, Bhat, Garibay, and Garibay}]{prottasha2025user}
Nusrat~Jahan Prottasha, Md~Kowsher, Hafijur Raman, Israt~Jahan Anny, Prakash Bhat, Ivan Garibay, and Ozlem Garibay. 2025.
\newblock User profile with large language models: Construction, updating, and benchmarking.
\newblock \emph{arXiv preprint arXiv:2502.10660}.

\bibitem[{Ren et~al.(2024)Ren, Wei, Xia, Su, Cheng, Wang, Yin, and Huang}]{ren2024representation}
Xubin Ren, Wei Wei, Lianghao Xia, Lixin Su, Suqi Cheng, Junfeng Wang, Dawei Yin, and Chao Huang. 2024.
\newblock Representation learning with large language models for recommendation.
\newblock In \emph{Proceedings of the ACM web conference 2024}, pages 3464--3475.

\bibitem[{Rendle et~al.(2012)Rendle, Freudenthaler, Gantner, and Schmidt-Thieme}]{rendle2012bpr}
Steffen Rendle, Christoph Freudenthaler, Zeno Gantner, and Lars Schmidt-Thieme. 2012.
\newblock Bpr: Bayesian personalized ranking from implicit feedback.
\newblock \emph{arXiv preprint arXiv:1205.2618}.

\bibitem[{Richardson et~al.(2023)Richardson, Zhang, Gillespie, Kar, Singh, Raeesy, Khan, and Sethy}]{richardson2023integrating}
Chris Richardson, Yao Zhang, Kellen Gillespie, Sudipta Kar, Arshdeep Singh, Zeynab Raeesy, Omar~Zia Khan, and Abhinav Sethy. 2023.
\newblock Integrating summarization and retrieval for enhanced personalization via large language models.
\newblock \emph{arXiv preprint arXiv:2310.20081}.

\bibitem[{Salemi et~al.(2024{\natexlab{a}})Salemi, Kallumadi, and Zamani}]{salemi2024optimization}
Alireza Salemi, Surya Kallumadi, and Hamed Zamani. 2024{\natexlab{a}}.
\newblock Optimization methods for personalizing large language models through retrieval augmentation.
\newblock In \emph{Proceedings of the 47th International ACM SIGIR Conference on Research and Development in Information Retrieval}, pages 752--762.

\bibitem[{Salemi et~al.(2024{\natexlab{b}})Salemi, Mysore, Bendersky, and Zamani}]{salemi2024lamp}
Alireza Salemi, Sheshera Mysore, Michael Bendersky, and Hamed Zamani. 2024{\natexlab{b}}.
\newblock Lamp: When large language models meet personalization.
\newblock In \emph{Proceedings of the 62nd Annual Meeting of the Association for Computational Linguistics (Volume 1: Long Papers)}, pages 7370--7392.

\bibitem[{Shao et~al.(2024)Shao, Wang, Zhu, Xu, Song, Bi, Zhang, Zhang, Li, Wu et~al.}]{shao2024deepseekmath}
Zhihong Shao, Peiyi Wang, Qihao Zhu, Runxin Xu, Junxiao Song, Xiao Bi, Haowei Zhang, Mingchuan Zhang, YK~Li, Yang Wu, and 1 others. 2024.
\newblock Deepseekmath: Pushing the limits of mathematical reasoning in open language models.
\newblock \emph{arXiv preprint arXiv:2402.03300}.

\bibitem[{Shi et~al.(2025)Shi, Xu, Zeqi, Zi, Wu, and Xu}]{shi2025personax}
Yunxiao Shi, Wujiang Xu, Zhang Zeqi, Xing Zi, Qiang Wu, and Min Xu. 2025.
\newblock Personax: A recommendation agent-oriented user modeling framework for long behavior sequence.
\newblock In \emph{Findings of the Association for Computational Linguistics: ACL 2025}, pages 5764--5787.

\bibitem[{Tan and Jiang(2023)}]{tan2023user}
Zhaoxuan Tan and Meng Jiang. 2023.
\newblock User modeling in the era of large language models: Current research and future directions.
\newblock \emph{arXiv preprint arXiv:2312.11518}.

\bibitem[{Tan et~al.(2024{\natexlab{a}})Tan, Liu, and Jiang}]{tan2024personalized}
Zhaoxuan Tan, Zheyuan Liu, and Meng Jiang. 2024{\natexlab{a}}.
\newblock Personalized pieces: Efficient personalized large language models through collaborative efforts.
\newblock In \emph{Proceedings of the 2024 Conference on Empirical Methods in Natural Language Processing}, pages 6459--6475.

\bibitem[{Tan et~al.(2024{\natexlab{b}})Tan, Zeng, Tian, Liu, Yin, and Jiang}]{tan2024democratizing}
Zhaoxuan Tan, Qingkai Zeng, Yijun Tian, Zheyuan Liu, Bing Yin, and Meng Jiang. 2024{\natexlab{b}}.
\newblock Democratizing large language models via personalized parameter-efficient fine-tuning.
\newblock In \emph{Proceedings of the 2024 Conference on Empirical Methods in Natural Language Processing}, pages 6476--6491.

\bibitem[{Tang et~al.(2024)Tang, Zhang, Ma, Lin, and T{\"u}re}]{tang2024found}
Raphael Tang, Crystina Zhang, Xueguang Ma, Jimmy Lin, and Ferhan T{\"u}re. 2024.
\newblock Found in the middle: Permutation self-consistency improves listwise ranking in large language models.
\newblock In \emph{Proceedings of the 2024 Conference of the North American Chapter of the Association for Computational Linguistics: Human Language Technologies (Volume 1: Long Papers)}, pages 2327--2340.

\bibitem[{Team et~al.(2023)Team, Anil, Borgeaud, Alayrac, Yu, Soricut, Schalkwyk, Dai, Hauth, Millican et~al.}]{team2023gemini}
Gemini Team, Rohan Anil, Sebastian Borgeaud, Jean-Baptiste Alayrac, Jiahui Yu, Radu Soricut, Johan Schalkwyk, Andrew~M Dai, Anja Hauth, Katie Millican, and 1 others. 2023.
\newblock Gemini: a family of highly capable multimodal models.
\newblock \emph{arXiv preprint arXiv:2312.11805}.

\bibitem[{Thoppilan et~al.(2022)Thoppilan, De~Freitas, Hall, Shazeer, Kulshreshtha, Cheng, Jin, Bos, Baker, Du et~al.}]{thoppilan2022lamda}
Romal Thoppilan, Daniel De~Freitas, Jamie Hall, Noam Shazeer, Apoorv Kulshreshtha, Heng-Tze Cheng, Alicia Jin, Taylor Bos, Leslie Baker, Yu~Du, and 1 others. 2022.
\newblock Lamda: Language models for dialog applications.
\newblock \emph{arXiv preprint arXiv:2201.08239}.

\bibitem[{Tseng et~al.(2024)Tseng, Huang, Hsiao, Chen, Huang, Meng, and Chen}]{tseng2024two}
Yu-Min Tseng, Yu-Chao Huang, Teng-Yun Hsiao, Wei-Lin Chen, Chao-Wei Huang, Yu~Meng, and Yun-Nung Chen. 2024.
\newblock Two tales of persona in llms: A survey of role-playing and personalization.
\newblock In \emph{Findings of the Association for Computational Linguistics: EMNLP 2024}, pages 16612--16631.

\bibitem[{Wang et~al.(2025)Wang, Zhang, Yang, Zhao, Liu, Zhan, Sun, Lin, Deng, Zhang et~al.}]{wang2025lettingo}
Lu~Wang, Di~Zhang, Fangkai Yang, Pu~Zhao, Jianfeng Liu, Yuefeng Zhan, Hao Sun, Qingwei Lin, Weiwei Deng, Dongmei Zhang, and 1 others. 2025.
\newblock Lettingo: Explore user profile generation for recommendation system.
\newblock In \emph{Proceedings of the 31st ACM SIGKDD Conference on Knowledge Discovery and Data Mining V. 2}, pages 2985--2995.

\bibitem[{Wang et~al.(2024)Wang, Li, Chen, Cai, Zhu, Lin, Cao, Kong, Liu, Liu et~al.}]{wang2024large}
Peiyi Wang, Lei Li, Liang Chen, Zefan Cai, Dawei Zhu, Binghuai Lin, Yunbo Cao, Lingpeng Kong, Qi~Liu, Tianyu Liu, and 1 others. 2024.
\newblock Large language models are not fair evaluators.
\newblock In \emph{Proceedings of the 62nd Annual Meeting of the Association for Computational Linguistics (Volume 1: Long Papers)}, pages 9440--9450.

\bibitem[{Wo{\'z}niak et~al.(2025)Wo{\'z}niak, Duszenko, Koco{\'n}, and Kazienko}]{wozniak2025improving}
Stanis{\l}aw Wo{\'z}niak, Jacek Duszenko, Jan Koco{\'n}, and Przemysaw Kazienko. 2025.
\newblock Improving llm-based recommender systems with user-controllable profiles.
\newblock In \emph{Companion Proceedings of the ACM on Web Conference 2025}, pages 2102--2111.

\bibitem[{Wu et~al.(2024)Wu, Shi, Rahmani, Ramineni, and Yilmaz}]{wu2024understanding}
Bin Wu, Zhengyan Shi, Hossein~A Rahmani, Varsha Ramineni, and Emine Yilmaz. 2024.
\newblock Understanding the role of user profile in the personalization of large language models.
\newblock \emph{arXiv preprint arXiv:2406.17803}.

\bibitem[{Yang et~al.(2025)Yang, Li, Yang, Zhang, Hui, Zheng, Yu, Gao, Huang, Lv et~al.}]{yang2025qwen3}
An~Yang, Anfeng Li, Baosong Yang, Beichen Zhang, Binyuan Hui, Bo~Zheng, Bowen Yu, Chang Gao, Chengen Huang, Chenxu Lv, and 1 others. 2025.
\newblock Qwen3 technical report.
\newblock \emph{arXiv preprint arXiv:2505.09388}.

\bibitem[{Zhang et~al.(2024)Zhang, Chen, Sheng, Wang, and Chua}]{zhang2024generative}
An~Zhang, Yuxin Chen, Leheng Sheng, Xiang Wang, and Tat-Seng Chua. 2024.
\newblock On generative agents in recommendation.
\newblock In \emph{Proceedings of the 47th international ACM SIGIR conference on research and development in Information Retrieval}, pages 1807--1817.

\bibitem[{Zhang et~al.(2025{\natexlab{a}})Zhang, Liu, Wang, Liu, Wu, Wang, and Chua}]{zhang2025personalized}
Jinghao Zhang, Yuting Liu, Wenjie Wang, Qiang Liu, Shu Wu, Liang Wang, and Tat-Seng Chua. 2025{\natexlab{a}}.
\newblock \href {https://doi.org/10.18653/v1/2025.acl-long.353} {Personalized text generation with contrastive activation steering}.
\newblock In \emph{Proceedings of the 63rd Annual Meeting of the Association for Computational Linguistics (Volume 1: Long Papers)}, pages 7128--7141, Vienna, Austria. Association for Computational Linguistics.

\bibitem[{Zhang et~al.(2026)Zhang, Xie, Hou, Zhao, Lin, and Wen}]{zhang2026recommendation}
Junjie Zhang, Ruobing Xie, Yupeng Hou, Wayne~Xin Zhao, Leyu Lin, and Ji-Rong Wen. 2026.
\newblock Recommendation as instruction following: A large language model empowered recommendation approach.
\newblock \emph{ACM Transactions on Information Systems}, 43(5):1--37.

\bibitem[{Zhang et~al.(2025{\natexlab{b}})Zhang, Wu, Zhou, and He}]{zhang2025proper}
Linhai Zhang, Jialong Wu, Deyu Zhou, and Yulan He. 2025{\natexlab{b}}.
\newblock \href {https://doi.org/10.18653/v1/2025.acl-long.800} {{PROPER}: A progressive learning framework for personalized large language models with group-level adaptation}.
\newblock In \emph{Proceedings of the 63rd Annual Meeting of the Association for Computational Linguistics (Volume 1: Long Papers)}, pages 16399--16411, Vienna, Austria. Association for Computational Linguistics.

\bibitem[{Zhang et~al.(2025{\natexlab{c}})Zhang, Adila, Shin, and Sala}]{zhang2025personalizellmfakealign}
Yijing Zhang, Dyah Adila, Changho Shin, and Frederic Sala. 2025{\natexlab{c}}.
\newblock \href {https://arxiv.org/abs/2503.01048} {Personalize your llm: Fake it then align it}.
\newblock \emph{Preprint}, arXiv:2503.01048.

\bibitem[{Zhuang et~al.(2024)Zhuang, Sun, Yu, Qiang, Wang, Zhang, and Dai}]{zhuang2024hydra}
Yuchen Zhuang, Haotian Sun, Yue Yu, Rushi Qiang, Qifan Wang, Chao Zhang, and Bo~Dai. 2024.
\newblock Hydra: Model factorization framework for black-box llm personalization.
\newblock \emph{Advances in Neural Information Processing Systems}, 37:100783--100815.

\bibitem[{Zollo et~al.(2025)Zollo, Siah, Ye, Li, and Namkoong}]{zollo2025personalllm}
Thomas Zollo, Andrew Siah, Naimeng Ye, Li~Li, and Hongseok Namkoong. 2025.
\newblock Personalllm: Tailoring llms to individual preferences.
\newblock In \emph{International Conference on Learning Representations}, volume 2025, pages 66949--66971.

\end{thebibliography}

\newpage
\appendix
\section{Appendix}
\begin{table*}[!htp]
\centering
\begin{tabular}{lccccc}
\toprule
Task   & Domain            & \#Train & \#Dev  & Avg.\ profile (train/dev) & Metric \\
\midrule
LaMP-1 & Citations         & 6{,}542  & 1{,}500 & 84.0 / 84.1   & Accuracy \\
LaMP-2 & Movie tags        & 5{,}073  & 1{,}410 & 101.6 / 65.5  & Accuracy, F1 \\
LaMP-3 & Product ratings   & 20{,}000 & 2{,}500 & 185.9 / 182.5 & MAE, RMSE      \\
LaMP-4 & News headlines    & 12{,}500 & 1{,}500 & 215.5 / 164.0 & ROUGE-L, ROGUE-1  \\
LaMP-5 & Paper titles      & 14{,}682 & 1{,}500 & 87.6 / 88.1   & ROUGE-L, ROGUE-1  \\
LaMP-7 & Tweet paraphrase  & 13{,}437 & 1{,}498 & 15.7 / 15.7   & ROUGE-L, ROGUE-1  \\
\bottomrule
\end{tabular}
\caption{LaMP statistics under the official time-based split. ``\#Train'' and ``\#Dev'' count unique users (each contributing one labeled query). ``Avg.\ profile'' is the mean number of historical interactions per user.}
\label{tab:lamp-stats}
\end{table*}
\subsection{Datasets}
\label{app:data}
We evaluate BUMP on the LaMP benchmark~\citep{salemi2024lamp}, the standard testbed for prompt-level LLM personalization. We cover six of the seven public LaMP tasks; LaMP-6 (personalized email writing) is omitted because its corpus is not publicly released. The six tasks together span five domains and three task types -- binary classification, multi-class classification, ordinal regression, and free-form generation:
\begin{itemize}[leftmargin=*]
  \item \textbf{LaMP-1: Personalized Citation Identification.} Binary classification of which of two candidate papers a user is more likely to cite, given their publication history.
  \item \textbf{LaMP-2: Personalized Movie Tagging.} 15-way classification of a movie tag (e.g.\ ``sci-fi'', ``dark comedy'') consistent with the user's prior tagging behavior; we use the updated ``LaMP-2 (new)'' split.
  \item \textbf{LaMP-3: Personalized Product Rating.} 1--5 ordinal rating prediction on Amazon reviews.
  \item \textbf{LaMP-4: Personalized News Headline Generation.} Generate a headline for a given article in the user's headline-writing style.
  \item \textbf{LaMP-5: Personalized Scholarly Title Generation.} Generate a paper title from an abstract in the user's authoring style.
  \item \textbf{LaMP-7: Personalized Tweet Paraphrasing.} Paraphrase a given tweet in the user's tweeting style.
\end{itemize}

\noindent\textbf{Statistics.}
\cref{tab:lamp-stats} reports the number of training and development users, the average number of historical interactions per user, and the official evaluation metric for each task. The six tasks cover a wide range of regimes: average profile lengths vary from $\sim$16 tweets (LaMP-7) to $>$180 reviews (LaMP-3), and training sizes range from $\sim$5K to 20K users.

\noindent\textbf{Training Split and Evaluation Protocol.}
For each dataset,we randomly sample 10\% of the training set as dev set for early stopping and hyper-parameter tuning and use the publicly released evaluation set as the testing set, as LaMP's testing sets are not publicly available. 
With respect to evaluation metrics, for LaMP-1, we use accuracy; for LaMP-2, we use accuracy as well as F1 score; for LaMP-3, we use MAE and RMSE ; and for LaMP-4, LaMP-5, and LaMP-7, we use ROGUE-1 and ROGUE-L.

\subsection{Prompts}
\label{app:prompt}
The
  forward and backward judge prompts of the BUMP reward are already
  shown in \cref{fig:forward-prompts} and \cref{fig:backward-prompts};
  below we provide (i) the profile-generation prompts consumed by
  $\pi_\theta$ at training and inference time
  (\cref{fig:grpo-summary-prompts}), (ii) the parallel prompts used by
  the zero-shot Gemini baselines (\cref{fig:gemini-summary-prompts}), and
  (iii) the downstream SFT input template that prepends the cached
  profile to every LaMP query (\cref{fig:sft-input-template}). All
  baselines that consume a textual user profile share the same SFT
  template, so the prefix string is the sole point of variation across
  methods.
    \begin{figure}[!htbp]
    \noindent\textbf{Forward / Backward judge prompts --- LaMP-2 (movie
    tagging).}
    \begin{promptblock}
    \small
    \textbf{Forward instruction:} You are an expert judge. Given a
    user summary, rank the following movies from most likely to least
    likely to have been tagged by this user.

    \vspace{0.4em}
    \textbf{Input --- User summary $s_u$:}
    ``A viewer who consistently tags character-driven indie dramas and
    90s thrillers; favors slow-burn pacing over spectacle.''

    \vspace{0.4em}
    \textbf{Input --- Movies to rank:}
    \begin{enumerate}[label=\textbf{[\arabic*]},leftmargin=2em,itemsep=0pt,align=left]
    \item Tag: \textit{thriller}; Title/Desc: \textit{Se7en} -- two
          detectives hunt a serial killer\ldots
    \item Tag: \textit{action}; Title/Desc: \textit{Fast \& Furious 9} --
          street racers reunite\ldots
    \item Tag: \textit{indie}; Title/Desc: \textit{Manchester by the Sea}
          -- a janitor returns home\ldots
    \item Tag: \textit{family}; Title/Desc: \textit{Paw Patrol Movie}\ldots
    \item Tag: \textit{drama}; Title/Desc: \textit{The Florida Project}\ldots
    \end{enumerate}
    \textbf{Output (judge):} \texttt{[3, 5, 1, 2, 4]}\\
    \textbf{Expected Groundtruth:} \texttt{[1, 3, 5, $\cdots$]}

    \vspace{0.6em}
    \textbf{Backward instruction:} You are an expert judge. Given a
    movie (with its tag), rank the following user summaries from most
    likely to least likely to describe its tagger.

    \vspace{0.4em}
    \textbf{Input --- Movie $h_u^{(p)}$:}
    Tag: \textit{indie}; Title/Desc: \textit{Manchester by the
    Sea} -- a janitor returns home after his brother's death\ldots

    \vspace{0.4em}
    \textbf{Input --- Summaries to rank:}
    \begin{enumerate}[label=\textbf{[\arabic*]},leftmargin=2em,itemsep=0pt,align=left]
    \item ``Action / blockbuster fan tagging mostly franchise tentpoles\ldots''
    \item ``Viewer who tags character-driven indie dramas and 90s thrillers\ldots''
    \item ``Animation enthusiast focused on family content\ldots''
    \item ``Horror completionist with a taste for 80s slashers\ldots''
    \end{enumerate}
    \textbf{Output (judge):} \texttt{[2, 1, 4, 3]} \\
    \textbf{Expected Groundtruth:} \texttt{[2, $\cdots$]}
    \end{promptblock}
    \vspace{-1em}
    \caption{LaMP-2 forward and backward judge prompts.}
    \label{fig:lamp2-prompts}
  \end{figure}

  \begin{figure}[!htbp]
    \noindent\textbf{Forward / Backward judge prompts --- LaMP-3 (product
    rating).}
    \begin{promptblock}
    \small
    \textbf{Forward instruction:} You are an expert judge. Given a
    user summary, rank the following product reviews from most likely to
    least likely to have been written by this user.

    \vspace{0.4em}
    \textbf{Input --- User summary $s_u$:}
    ``A pragmatic reviewer of kitchen and small-appliance products;
    rates strictly on durability and value; tends to give 4 stars by
    default, dropping to 2 only after material defects.''

    \vspace{0.4em}
    \textbf{Input --- Reviews to rank:}
    \begin{enumerate}[label=\textbf{[\arabic*]},leftmargin=2em,itemsep=0pt,align=left]
    \item Rating 4/5: ``Solid blender, gasket feels cheap but it works\ldots''
    \item Rating 5/5: ``OMG best mascara ever!!! holy grail\ldots''
    \item Rating 2/5: ``Toaster died after 6 weeks, returned\ldots''
    \item Rating 5/5: ``Loved this thriller, couldn't put it down\ldots''
    \item Rating 4/5: ``Kettle boils fast, lid hinge already loose\ldots''
    \end{enumerate}
    \textbf{Output (judge):} \texttt{[1, 5, 3, 2, 4]}\\
    \textbf{Expected Groundtruth:} \texttt{[1, 3, 5, $\cdots$]}

    \vspace{0.6em}
    \textbf{Backward instruction:} You are an expert judge. Given a
    product review (with its rating), rank the following user summaries
    from most likely to least likely to describe its author.

    \vspace{0.4em}
    \textbf{Input --- Review $h_u^{(p)}$:}
    Rating 2/5: ``Toaster died after 6 weeks, returned. Decent until
    then but I can't recommend something with this failure rate\ldots''

    \vspace{0.4em}
    \textbf{Input --- Summaries to rank:}
    \begin{enumerate}[label=\textbf{[\arabic*]},leftmargin=2em,itemsep=0pt,align=left]
    \item ``Enthusiastic 5-star beauty product reviewer\ldots''
    \item ``Pragmatic small-appliance reviewer rating on durability\ldots''
    \item ``Sci-fi book reader leaving long narrative reviews\ldots''
    \item ``Brand-loyal tech reviewer; rarely below 4 stars\ldots''
    \end{enumerate}
    \textbf{Output (judge):} \texttt{[2, 4, 3, 1]} \\
    \textbf{Expected Groundtruth:} \texttt{[2, $\cdots$]}
    \end{promptblock}
    \vspace{-1em}
    \caption{LaMP-3 forward and backward judge prompts.}
    \label{fig:lamp3-prompts}
  \end{figure}

  \begin{figure}[!htbp]
    \noindent\textbf{Forward / Backward judge prompts --- LaMP-4 (news
    headline generation).}
    \begin{promptblock}
    \small
    \textbf{Forward instruction:} You are an expert judge. Given a
    user summary, rank the following news articles from most likely to
    least likely to have been headlined by this user.

    \vspace{0.4em}
    \textbf{Input --- User summary $s_u$:}
    ``A lifestyle and wellness writer who favors punchy, second-person
    headlines centered on relationships, sleep, and self-care; avoids
    hard-news political framing.''

    \vspace{0.4em}
    \textbf{Input --- Articles to rank:}
    \begin{enumerate}[label=\textbf{[\arabic*]},leftmargin=2em,itemsep=0pt,align=left]
    \item Headline: ``Why You Keep Waking Up at 3 AM''; Article: sleep
          researcher Q\&A on cortisol cycles\ldots
    \item Headline: ``Senate Passes Defense Authorization Bill''\ldots
    \item Headline: ``5 Tiny Habits That Saved My Marriage''\ldots
    \item Headline: ``Quarterly Earnings: TSMC Beats Estimates''\ldots
    \item Headline: ``Your Morning Coffee Might Be Lying to You''\ldots
    \end{enumerate}
    \textbf{Output (judge):} \texttt{[1, 3, 5, 4, 2]}\\
    \textbf{Expected Groundtruth:} \texttt{[1, 3, 5, $\cdots$]}

    \vspace{0.6em}
    \textbf{Backward instruction:} You are an expert judge. Given a
    news article, rank the following user summaries from most likely to
    least likely to describe its headline writer.

    \vspace{0.4em}
    \textbf{Input --- Article $h_u^{(p)}$:}
    Headline: ``5 Tiny Habits That Saved My Marriage''; Article: a
    personal essay on routine, communication, and small daily
    rituals\ldots

    \vspace{0.4em}
    \textbf{Input --- Summaries to rank:}
    \begin{enumerate}[label=\textbf{[\arabic*]},leftmargin=2em,itemsep=0pt,align=left]
    \item ``Hard-news political reporter; declarative, neutral
          headlines\ldots''
    \item ``Lifestyle and wellness writer with punchy second-person
          headlines\ldots''
    \item ``Financial markets writer; numeric, ticker-heavy
          headlines\ldots''
    \item ``Sports columnist; rhetorical-question headlines about teams\ldots''
    \end{enumerate}
    \textbf{Output (judge):} \texttt{[2, 4, 1, 3]} \\
    \textbf{Expected Groundtruth:} \texttt{[2, $\cdots$]}
    \end{promptblock}
    \vspace{-1em}
    \caption{LaMP-4 forward and backward judge prompts.}
    \label{fig:lamp4-prompts}
  \end{figure}

  \begin{figure}[!htbp]
    \noindent\textbf{Forward / Backward judge prompts --- LaMP-5
    (scholarly title generation).}
    \begin{promptblock}
    \small
    \textbf{Forward instruction:} You are an expert judge. Given a
    user summary, rank the following academic works (abstract + title)
    from most likely to least likely to have been titled by this user.

    \vspace{0.4em}
    \textbf{Input --- User summary $s_u$:}
    ``A graph machine-learning researcher whose titles favor
    ``X for Y''-style framings emphasizing scalability and
    representation learning on large heterogeneous graphs.''

    \vspace{0.4em}
    \textbf{Input --- Works to rank:}
    \begin{enumerate}[label=\textbf{[\arabic*]},leftmargin=2em,itemsep=0pt,align=left]
    \item Title: ``Scalable Heterogeneous GNNs for Web-Scale
          Recommendation''; Abstract: \ldots
    \item Title: ``A Randomized Controlled Trial of\ldots'';
          Abstract: clinical trial methodology\ldots
    \item Title: ``Subgraph Sampling for Trillion-Edge Representation
          Learning''; Abstract: \ldots
    \item Title: ``On the Stability of Tokamak Plasma\ldots''; Abstract:
          \ldots
    \item Title: ``Graph Attention for Cross-Domain Cold Start''; Abstract:
          \ldots
    \end{enumerate}
    \textbf{Output (judge):} \texttt{[1, 3, 5, 4, 2]}\\
    \textbf{Expected Groundtruth:} \texttt{[1, 3, 5, $\cdots$]}

    \vspace{0.6em}
    \textbf{Backward instruction:} You are an expert judge. Given an
    academic work (abstract + title), rank the following user summaries
    from most likely to least likely to describe its author.

    \vspace{0.4em}
    \textbf{Input --- Work $h_u^{(p)}$:}
    Title: ``Subgraph Sampling for Trillion-Edge Representation Learning'';
    Abstract: we propose a sampling scheme that scales GNN training to
    industrial graphs\ldots

    \vspace{0.4em}
    \textbf{Input --- Summaries to rank:}
    \begin{enumerate}[label=\textbf{[\arabic*]},leftmargin=2em,itemsep=0pt,align=left]
    \item ``Clinical-trial methodologist; titles emphasize study design\ldots''
    \item ``Graph ML researcher favoring `X for Y' titles on scalable
          representation learning\ldots''
    \item ``Plasma physicist; titles foreground stability and confinement\ldots''
    \item ``NLP researcher focused on instruction tuning and alignment\ldots''
    \end{enumerate}
    \textbf{Output (judge):} \texttt{[2, 4, 1, 3]} \\
    \textbf{Expected Groundtruth:} \texttt{[2, $\cdots$]}
    \end{promptblock}
    \vspace{-1em}
    \caption{LaMP-5 forward and backward judge prompts.}
    \label{fig:lamp5-prompts}
  \end{figure}

  \begin{figure}[!htbp]
    \noindent\textbf{Forward / Backward judge prompts --- LaMP-7 (tweet
    paraphrasing).}
    \begin{promptblock}
    \small
    \textbf{Forward instruction:} You are an expert judge. Given a
    user summary, rank the following tweets from most likely to least
    likely to have been written by this user.

    \vspace{0.4em}
    \textbf{Input --- User summary $s_u$:}
    ``A casual, lowercase-only tweeter who posts mostly about NBA games
    and late-night food; uses dry self-deprecating humor and rarely
    hashtags.''

    \vspace{0.4em}
    \textbf{Input --- Tweets to rank:}
    \begin{enumerate}[label=\textbf{[\arabic*]},leftmargin=2em,itemsep=0pt,align=left]
    \item ``lakers in 6. cope.''
    \item ``Excited to announce my new role @MegaCorp! \#blessed
          \#newchapter''
    \item ``ate cold pizza standing up at 1am, peak fine dining''
    \item ``URGENT: Sign the petition NOW \#ActOnClimate''
    \item ``halftime score is rigged i cant do this anymore''
    \end{enumerate}
    \textbf{Output (judge):} \texttt{[1, 3, 5, 2, 4]}\\
    \textbf{Expected Groundtruth:} \texttt{[1, 3, 5, $\cdots$]}

    \vspace{0.6em}
    \textbf{Backward instruction:} You are an expert judge. Given a
    tweet, rank the following user summaries from most likely to least
    likely to describe its author.

    \vspace{0.4em}
    \textbf{Input --- Tweet $h_u^{(p)}$:}
    ``ate cold pizza standing up at 1am, peak fine dining''

    \vspace{0.4em}
    \textbf{Input --- Summaries to rank:}
    \begin{enumerate}[label=\textbf{[\arabic*]},leftmargin=2em,itemsep=0pt,align=left]
    \item ``Corporate-comms-style tweeter; capitalized, hashtagged
          announcements\ldots''
    \item ``Casual lowercase NBA + late-night-food tweeter with dry
          self-deprecating humor\ldots''
    \item ``Activist account; urgent calls-to-action with petitions and
          hashtags\ldots''
    \item ``Tech-industry thread writer; numbered lists and product
          takes\ldots''
    \end{enumerate}
    \textbf{Output (judge):} \texttt{[2, 4, 1, 3]} \\
    \textbf{Expected Groundtruth:} \texttt{[2, $\cdots$]}
    \end{promptblock}
    \vspace{-1em}
    \caption{LaMP-7 forward and backward judge prompts.}
    \label{fig:lamp7-prompts}
  \end{figure}
  
  \begin{figure}[!htbp]
    \noindent\textbf{Summary-generation prompt for BUMP --- instruction
    given to $\pi_\theta$ to elicit the profile $s_u$ from a user's
    visible history $H_u^{\mathrm{vis}}$.}
    \begin{promptblock}
    \small
    \textbf{Instruction (LaMP-1, citation identification):}
    Below is a list of academic works authored by a user:

    \texttt{\{profile\}}

    Based on these works, please write a concise summary about the user.

    \vspace{0.5em}
    \textbf{Instruction (LaMP-2, movie tagging):}
    Below is a list of movies previously tagged by a user:

    \texttt{\{profile\}}

    Based on these tags and descriptions, please write a concise summary
    about the user's movie preferences and tagging patterns.

    \vspace{0.5em}
    \textbf{Instruction (LaMP-3, product rating):}
    Below is a list of product reviews previously written by a user,
    each with the rating they gave:

    \texttt{\{profile\}}

    Based on these reviews and ratings, please write a concise summary
    describing the user's rating patterns, preferences, and review style.

    \vspace{0.5em}
    \textbf{Instruction (LaMP-4, news headline generation):}
    Below is a list of news articles previously titled by a user:

    \texttt{\{profile\}}

    Based on these articles and their titles, please write a concise
    summary describing the user's writing style, tone, and headline
    preferences.

    \vspace{0.5em}
    \textbf{Instruction (LaMP-5, scholarly title generation):}
    Below is a list of academic works (title + abstract) previously
    titled by a user:

    \texttt{\{profile\}}

    Based on these works, please write a concise summary describing the
    user's research interests, themes, and title-writing style.

    \vspace{0.5em}
    \textbf{Instruction (LaMP-7, tweet paraphrasing):}
    Below is a list of tweets previously written by a user:

    \texttt{\{profile\}}

    Based on these tweets, please write a concise summary describing the
    user's tweeting style, tone, typical topics, and personality.

    \vspace{0.5em}
    \texttt{\{profile\}} is the user's visible history rendered with a
    per-task item template, e.g.\ ``\texttt{- Title: \{title\}\textbackslash n
    Abstract: \{abstract\}}'' for LaMP-1, ``\texttt{- [\{tag\}]
    \{description\}}'' for LaMP-2, and ``\texttt{- Review (rated
    \{score\}/5): \{text\}}'' for LaMP-3.
    \end{promptblock}
    \caption{Profile-generation prompts used by BUMP, BUMP+, the
    Direct-Reward baseline, and the zero-shot Qwen3-4B profile baseline.
    All four methods share the exact same string; they differ only in
    what model emits the profile and under what reward.}
    \label{fig:grpo-summary-prompts}
  \end{figure}

  \begin{figure}[!htbp]
    \noindent\textbf{Profile-generation prompt for the Gemini zero-shot
    baselines (\textsc{Gemini-3-Flash}, \textsc{Gemini-3-Pro}).}
    \begin{promptblock}
    \small
    \textbf{Instruction (LaMP-1):}
    Below is a list of works authored by a user:

    \texttt{\{user\_history\}}

    Please write a concise summary (2--3 sentences) about the user.

    \vspace{0.5em}
    \textbf{Instruction (LaMP-2):}
    Below is a list of movies previously tagged by a user:

    \texttt{\{user\_history\}}

    Please write a concise summary (2--3 sentences) about the user's
    movie preferences and tagging patterns.

    \vspace{0.5em}
    \textbf{Instruction (LaMP-3):}
    Below is a list of product reviews previously written by a user, each
    with the rating they gave:

    \texttt{\{user\_history\}}

    Please write a concise summary (2--3 sentences) about the user's
    rating patterns, preferences, and review style.

    \vspace{0.5em}
    \textbf{Instruction (LaMP-4):}
    Below is a list of news articles previously titled by a user:

    \texttt{\{user\_history\}}

    Please write a concise summary (2--3 sentences) about the user's
    writing style, tone, and headline preferences.

    \vspace{0.5em}
    \textbf{Instruction (LaMP-5):}
    Below is a list of academic works (title + abstract) previously
    titled by a user:

    \texttt{\{user\_history\}}

    Please write a concise summary (2--3 sentences) about the user's
    research interests, themes, and title-writing style.

    \vspace{0.5em}
    \textbf{Instruction (LaMP-7):}
    Below is a list of tweets previously written by a user:

    \texttt{\{user\_history\}}

    Please write a concise summary (2--3 sentences) about the user's
    tweeting style, tone, typical topics, and personality.

    \vspace{0.5em}
    \texttt{\{user\_history\}} is rendered with the same per-task item
    templates as \cref{fig:grpo-summary-prompts}.
    \end{promptblock}
    \caption{Zero-shot Gemini profile-generation prompts. The only
    systematic difference from the BUMP prompt of
    \cref{fig:grpo-summary-prompts} is the explicit ``(2--3 sentences)''
    length constraint, which substitutes for the soft length penalty
    $R_{\mathrm{len}}$ that BUMP applies during GRPO --- the Gemini API
    baselines are not trained, so no learned length control is available.}
    \label{fig:gemini-summary-prompts}
  \end{figure}

  \begin{figure}[!htbp]
    \noindent\textbf{Downstream SFT input template --- prepends the cached
    profile to every LaMP query during LoRA-SFT.}
    \begin{promptblock}
    \small
    \textbf{Input:}
    This is a summary about this user: \texttt{\{summary\}}

    \texttt{\{input\}}
    \end{promptblock}
    \caption{LoRA-SFT input template. \texttt{\{summary\}} is the cached
    profile and \texttt{\{input\}} is the LaMP question (unchanged
    from the official LaMP release). Shared across every profile-based
    method in \cref{tab:lamp-main}, so the only point of variation is the
    \texttt{\{summary\}} string itself.}
    \label{fig:sft-input-template}
  \end{figure}
  
\subsection{Direct Reward Formulation}
\label{app:direct_reward}
The \textsc{Direct Reward} baseline (\cref{tab:lamp-main}) replaces our bidirectional NDCG reward with a downstream LaMP-task signal while keeping every other component fixed: identical policy (\textsc{Qwen3-4B-Instruct}), frozen judge $J$, length penalty $R_{\mathrm{len}}$, and GRPO hyperparameters. This isolates the contribution of self-supervision from the contribution of GRPO itself.
For each rollout the judge is conditioned on the candidate profile $s_u$ \emph{together with} the LaMP question $q_u$, and is asked to solve the downstream task directly from $s_u$. The per-rollout reward is task-specific:

\noindent\textbf{Classification (LaMP-1, LaMP-2).} \\
The judge predicts one of $L$ allowed labels --- the two candidate papers for LaMP-1, the 15 movie tags for LaMP-2 --- with the output space enforced by vLLM's \texttt{guided\_choice} constrained decoding to eliminate hallucination. 
The reward is exact match:
\begin{equation*}
R_{\mathrm{cls}}(s_u) = \mathbb{1}\!\left[J(s_u, q_u) = y_u\right],
\end{equation*}
where $y_u$ is the gold label.

\noindent\textbf{Ordinal regression (LaMP-3).} \\
The judge predicts a rating $\hat{y}\in\{1,\dots,5\}$, again under guided decoding. The reward is graded by distance to the gold rating $y_u$:
\begin{equation*}
R_{\mathrm{ord}}(s_u) = \max\!\Big(0,\;
 1 - \tfrac{|J(s_u, q_u) - y_u|}{4}\Big).
\end{equation*}

\noindent\textbf{Generation (LaMP-4, LaMP-5, LaMP-7).} \\
The judge freely decodes the target text (a headline / paper title / tweet paraphrase). The reward is ROUGE-L F1 against the gold target:
\begin{equation*}
R_{\mathrm{gen}}(s_u)
= \mathrm{ROUGE\text{-}L}\!\left(J(s_u, q_u),\; y_u\right).
\end{equation*}

In all three cases the task reward is combined with the same length penalty used by BUMP, yielding $R(s_u) = R_{\mathrm{task}}(s_u) - R_{\mathrm{len}}(s_u)$, and is optimized with GRPO under the same group size $G{=}8$ and per-device batch size as BUMP. Every other knob (epochs, learning rate, threshold $T$) is identical, so the only point of variation between \textsc{Direct Reward} and BUMP is the reward function itself.
\subsection{Perf. w.r.t. Different Judges and Backbones}
\label{app:backbone}
\cref{tab:ablation-judge} and \cref{tab:ablation-policy} sweep the size
  of the frozen judge $J$ and the GRPO policy $\pi_\theta$ independently
  over $\{0.6\text{B}, 4\text{B}, 8\text{B}\}$ Qwen3-Instruct variants. In
  both cases the 4B configuration is the default and matches the BUMP
  row of \cref{tab:lamp-main}; we vary one component and hold the other
  at 4B.

  \begin{table}[!htbp]
    \centering
    \small
    \setlength{\tabcolsep}{4pt}
    \renewcommand{\arraystretch}{1.1}
    \begin{tabular}{l l c c c}
    \toprule
    Task & Metric
      & \shortstack{Judge\\0.6B}
      & \shortstack{Judge\\4B}
      & \shortstack{Judge\\8B} \\
    \midrule
    LaMP-1 & Acc ($\uparrow$)        & 77.5  & 80.9 & 81.0 \\
    \midrule
    \multirow{2}{*}{LaMP-2}
           & Acc ($\uparrow$)        & 87.5  & 89.4 & 89.5 \\
           & F1 ($\uparrow$)         & 82.5  & 84.5 & 84.5 \\
    \midrule
    \multirow{2}{*}{LaMP-3}
           & MAE ($\downarrow$)      & 0.232 & 0.226 & 0.225 \\
           & RMSE ($\downarrow$)     & 0.529 & 0.523 & 0.522 \\
    \midrule
    \multirow{2}{*}{LaMP-4}
           & ROUGE-1 ($\uparrow$)    & 0.190 & 0.197 & 0.197 \\
           & ROUGE-L ($\uparrow$)    & 0.172 & 0.179 & 0.179 \\
    \midrule
    \multirow{2}{*}{LaMP-5}
           & ROUGE-1 ($\uparrow$)    & 0.484 & 0.490 & 0.491 \\
           & ROUGE-L ($\uparrow$)    & 0.434 & 0.439 & 0.440 \\
    \midrule
    \multirow{2}{*}{LaMP-7}
           & ROUGE-1 ($\uparrow$)    & 0.524 & 0.530 & 0.531 \\
           & ROUGE-L ($\uparrow$)    & 0.471 & 0.477 & 0.477 \\
    \bottomrule
    \end{tabular}
    \caption{Judge-size sweep. Policy $\pi_\theta$ fixed at 4B.}
    \label{tab:ablation-judge}
  \end{table}

  \begin{table}[!htbp]
    \centering
    \small
    \setlength{\tabcolsep}{4pt}
    \renewcommand{\arraystretch}{1.1}
    \begin{tabular}{l l c c c}
    \toprule
    Task & Metric
      & \shortstack{Policy\\0.6B}
      & \shortstack{Policy\\4B}
      & \shortstack{Policy\\8B} \\
    \midrule
    LaMP-1 & Acc ($\uparrow$)        & 70.2  & 80.9 & 82.5 \\
    \midrule
    \multirow{2}{*}{LaMP-2}
           & Acc ($\uparrow$)        & 82.5  & 89.4 & 90.2 \\
           & F1 ($\uparrow$)         & 76.5  & 84.5 & 85.4 \\
    \midrule
    \multirow{2}{*}{LaMP-3}
           & MAE ($\downarrow$)      & 0.245 & 0.226 & 0.219 \\
           & RMSE ($\downarrow$)     & 0.545 & 0.523 & 0.518 \\
    \midrule
    \multirow{2}{*}{LaMP-4}
           & ROUGE-1 ($\uparrow$)    & 0.178 & 0.197 & 0.203 \\
           & ROUGE-L ($\uparrow$)    & 0.160 & 0.179 & 0.184 \\
    \midrule
    \multirow{2}{*}{LaMP-5}
           & ROUGE-1 ($\uparrow$)    & 0.470 & 0.490 & 0.498 \\
           & ROUGE-L ($\uparrow$)    & 0.420 & 0.439 & 0.445 \\
    \midrule
    \multirow{2}{*}{LaMP-7}
           & ROUGE-1 ($\uparrow$)    & 0.510 & 0.530 & 0.537 \\
           & ROUGE-L ($\uparrow$)    & 0.460 & 0.477 & 0.483 \\
    \bottomrule
    \end{tabular}
    \caption{Policy-size sweep. Judge $J$ fixed at 4B.}
    \label{tab:ablation-policy}
  \end{table}

\noindent\textbf{Judge size saturates at 4B.} Scaling the judge from 0.6B to 4B yields the bulk of the gain on every task (e.g.,\ LaMP-1 $77.5\!\to\!80.9$, LaMP-3 MAE $0.232\!\to\!0.226$): a sub-1B judge produces noisy rankings that propagate as noisy gradients. Pushing further to 8B brings no consistent improvement --- differences fall within seed-level noise on all $11$ rows. The judge's task (list-ranking $1+K$ candidates) appears bounded in difficulty: once $J$ is capable of producing low-variance rankings, additional scale is wasted.

\noindent\textbf{Policy size keeps scaling.} The GRPO policy, in contrast, benefits from every increment. The 0.6B policy is sharply worse across the board (e.g.,\ LaMP-2 F1 drops $84.5\!\to\!76.5$), consistent with $\pi_\theta$ needing enough capacity to emit a coherent, on-distribution profile. Scaling to 8B yields a further consistent lift over the 4B default (e.g.,\ LaMP-1 $80.9\!\to\!82.5$, LaMP-4 ROUGE-L $0.179\!\to\!0.184$). Profile generation is the harder, less bounded side of the system, and self-supervision evidently scales with policy capacity --- a useful property given that the cost of serving a larger \emph{judge} is what otherwise limits joint scaling.
\subsection{Reward Trajectory of BUMP}
\label{app:reward_curve}
  \begin{figure*}[!htbp]
    \centering
    \includegraphics[width=\linewidth]{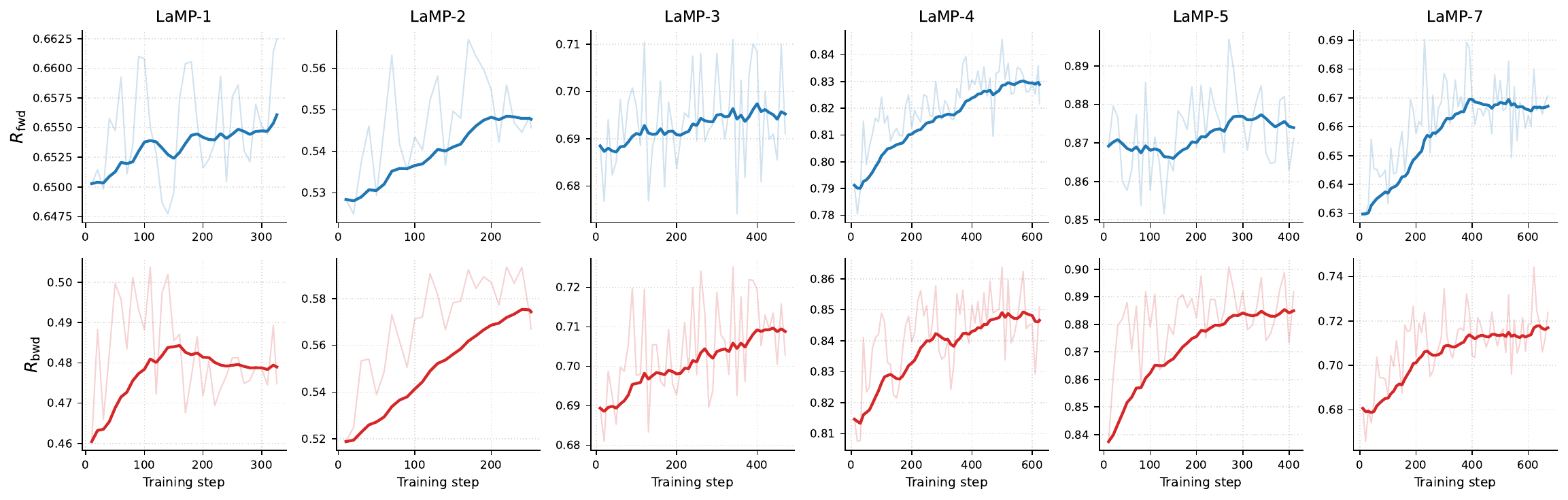}
    \caption{Training dynamics of the bidirectional reward across the six
  LaMP tasks. Each column shows one task; the top row plots
  $R_{\mathrm{fwd}}$ (blue) and the bottom row plots $R_{\mathrm{bwd}}$
  (red). Light traces are the raw per-step mean; bold curves apply an
  EWMA smoother ($\alpha{=}0.1$). Y-axes are scaled independently per
  subplot to highlight relative motion. Both directions rise
  monotonically on every task, confirming
  that the bidirectional reward delivers usable gradient throughout
  training.}
    \label{fig:reward-curves}
  \end{figure*}

\cref{fig:reward-curves} plots the per-step mean of the two reward
  components across the six LaMP tasks. Two patterns hold uniformly.

  \noindent\textbf{Both directions improve monotonically on every task.}
  Across all six tasks, both $R_{\mathrm{fwd}}$ and $R_{\mathrm{bwd}}$
  rise smoothly throughout training, including LaMP-1 (
  $R_{\mathrm{fwd}}\!:\;0.65\!\to\!0.66$,
  $R_{\mathrm{bwd}}\!:\;0.45\!\to\!0.50$), where the absolute range
  is small but the trend is clean. The bidirectional reward thus
  delivers usable gradient over the full training trajectory rather than
  collapsing once one direction saturates.

  \noindent\textbf{The backward term carries more of the visible motion.}
  On every task, $\Delta R_{\mathrm{bwd}}$ exceeds $\Delta
  R_{\mathrm{fwd}}$, often by a factor of two or more (e.g.\ LaMP-7:
  $\Delta R_{\mathrm{bwd}} \approx 0.07$ vs.\ $\Delta R_{\mathrm{fwd}}
  \approx 0.03$; LaMP-1: $\Delta R_{\mathrm{bwd}} \approx 0.05$ vs.\
  $\Delta R_{\mathrm{fwd}} \approx 0.01$). $R_{\mathrm{fwd}}$ starts
  close to ceiling --- LaMP histories carry strong topical signal that
  even a zero-shot profile already preserves, so the forward ranking
  task is already partially solved at step 0. $R_{\mathrm{bwd}}$ instead
  requires the profile to be \emph{uniquely identifying} among other
  users' profiles, a harder property that the policy must actively
  learn, which is where most of the visible training-time progress
  shows up.
\subsection{Analysis on Debiasing the Judge}
\label{app:judge_debias}
\begin{table}[!htbp]
    \centering
    \small
    \setlength{\tabcolsep}{4pt}
    \renewcommand{\arraystretch}{1.1}
    \begin{tabular}{l l c c c c}
    \toprule
    Task & Metric
      & $M{=}1$
      & $M{=}2$
      & $M{=}3$
      & $M{=}4$ \\
    \midrule
    LaMP-1 & Acc ($\uparrow$)        & 79.0  & 80.9 & 81.0 & 81.0 \\
    \midrule
    \multirow{2}{*}{LaMP-2}
           & Acc ($\uparrow$)        & 88.0  & 89.4 & 89.5 & 89.5 \\
           & F1 ($\uparrow$)         & 82.8  & 84.5 & 84.5 & 84.6 \\
    \midrule
    \multirow{2}{*}{LaMP-3}
           & MAE ($\downarrow$)      & 0.232 & 0.226 & 0.226 & 0.225 \\
           & RMSE ($\downarrow$)     & 0.529 & 0.523 & 0.523 & 0.522 \\
    \midrule
    \multirow{2}{*}{LaMP-4}
           & ROUGE-1 ($\uparrow$)    & 0.190 & 0.197 & 0.197 & 0.197 \\
           & ROUGE-L ($\uparrow$)    & 0.172 & 0.179 & 0.180 & 0.180 \\
    \midrule
    \multirow{2}{*}{LaMP-5}
           & ROUGE-1 ($\uparrow$)    & 0.483 & 0.490 & 0.491 & 0.491 \\
           & ROUGE-L ($\uparrow$)    & 0.432 & 0.439 & 0.440 & 0.440 \\
    \midrule
    \multirow{2}{*}{LaMP-7}
           & ROUGE-1 ($\uparrow$)    & 0.523 & 0.530 & 0.530 & 0.531 \\
           & ROUGE-L ($\uparrow$)    & 0.470 & 0.477 & 0.477 & 0.478 \\
    \bottomrule
    \end{tabular}
    \caption{Ablation on the number of debiasing permutations $M$. All
    other settings match \cref{tab:lamp-main}; $M{=}2$ is the default.}
    \label{tab:ablation-debias}
  \end{table}

  \cref{tab:ablation-debias} sweeps the number of debiasing permutations
  $M\in\{1,2,3,4\}$. At $M{=}1$ the per-rollout reward is computed from
  a single arbitrary candidate ordering and inherits the judge's
  position bias directly; performance drops on every row (e.g.\ LaMP-1
  $80.9\!\to\!79.0$, LaMP-2 F1 $84.5\!\to\!82.8$, LaMP-4 ROUGE-L
  $0.179\!\to\!0.172$). $M{=}2$ already captures the bulk of the
  benefit --- the variance of the Monte-Carlo NDCG estimator falls as
  $1/M$, so two permutations halve the dominant noise term --- and
  further increases to $M{=}3$ or $M{=}4$ yield only seed-level
  differences. Since each additional permutation costs one extra judge
  call per rollout, $M{=}2$ is the natural operating point and matches
  the default used in \cref{tab:lamp-main}.
\subsection{Complementarity of BUMP}
\label{app:complement}
Self-supervised and task-supervised rewards capture different aspects
  of profile quality: the bidirectional NDCG reward of BUMP optimizes
  the profile to be a faithful, identity-preserving description of the
  user, while the Direct Reward of \cref{app:direct_reward} optimizes it
  to maximize a single labeled task signal. These objectives are not
  redundant. To test whether they combine cleanly, we train a third
  variant that augments BUMP+'s reward with the per-task Direct Reward
  at unit weight, yielding the per-rollout reward
  \begin{equation*}
  R(s_u)
  = \bar{R}^{u}_{\mathrm{fwd}}
  + \bar{R}^{u}_{\mathrm{bwd}}
  + R_{\mathrm{task}}(s_u)
  - R_{\mathrm{len}}(s_u),
  \end{equation*}
  optimized with the same GRPO settings as the rest of the paper.

  \begin{table}[!htbp]
    \centering
    \small
    \setlength{\tabcolsep}{4pt}
    \renewcommand{\arraystretch}{1.1}
    \begin{tabular}{l l c c c}
    \toprule
    Task & Metric
      & \shortstack{Direct\\Reward}
      & \shortstack{BUMP+}
      & \shortstack{BUMP+\\+ Direct} \\
    \midrule
    LaMP-1 & Acc ($\uparrow$)        & 79.3  & 80.1 & \textbf{81.0} \\
    \midrule
    \multirow{2}{*}{LaMP-2}
           & Acc ($\uparrow$)        & 88.9  & 89.9 & \textbf{90.3} \\
           & F1 ($\uparrow$)         & 84.0  & 84.7 & \textbf{85.0} \\
    \midrule
    \multirow{2}{*}{LaMP-3}
           & MAE ($\downarrow$)      & 0.218 & 0.214 & \textbf{0.211} \\
           & RMSE ($\downarrow$)     & 0.517 & 0.517 & \textbf{0.514} \\
    \midrule
    \multirow{2}{*}{LaMP-4}
           & ROUGE-1 ($\uparrow$)    & 0.200 & 0.199 & \textbf{0.205} \\
           & ROUGE-L ($\uparrow$)    & 0.182 & 0.180 & \textbf{0.186} \\
    \midrule
    \multirow{2}{*}{LaMP-5}
           & ROUGE-1 ($\uparrow$)    & 0.491 & 0.494 & \textbf{0.498} \\
           & ROUGE-L ($\uparrow$)    & 0.442 & 0.443 & \textbf{0.447} \\
    \midrule
    \multirow{2}{*}{LaMP-7}
           & ROUGE-1 ($\uparrow$)    & 0.533 & 0.536 & \textbf{0.540} \\
           & ROUGE-L ($\uparrow$)    & 0.479 & 0.482 & \textbf{0.486} \\
    \bottomrule
    \end{tabular}
    \caption{Complementarity of self-supervised and task-supervised
    rewards. The combined reward (right column) wins on every row.
    Best per-row result in \textbf{bold}.}
    \label{tab:ablation-complement}
  \end{table}

  \noindent\textbf{Combined reward wins on every row.}
  \cref{tab:ablation-complement} shows that the combined variant matches
  or beats both parents on all $11$ metric rows (e.g.\ LaMP-1
  $81.0$ vs.\ $80.1\,/\,79.3$; LaMP-2 F1 $85.0$ vs.\ $84.7\,/\,84.0$;
  LaMP-3 MAE $0.211$ vs.\ $0.214\,/\,0.218$; LaMP-4 ROUGE-L
  $0.186$ vs.\ $0.180\,/\,0.182$). The two reward formulations capture
  complementary signal: the self-supervised term enforces broad user
  fidelity that transfers across tasks, while the task-supervised term
  sharpens the profile toward whatever idiosyncrasy the current LaMP
  task happens to need.

  \noindent\textbf{Practical implication.} Our central thesis is that
  self-supervision is a viable \emph{replacement} for downstream labels
  (\cref{tab:lamp-main}). This experiment shows the stronger
  complementary claim: when labels \emph{are} available, BUMP can be
  layered on top of the task reward at no architectural cost and acts as
  a strict regularizer, rather than a competing alternative.

\subsection{Additional Hyperparameters}
  \label{app:hyperparams}
  We report the hyperparameters used by the two training stages in our
  pipeline: GRPO for the profile generator $\pi_\theta$
  (\cref{sec:method:opt}) and LoRA-SFT for the downstream backbone that
  consumes the cached profile (\cref{sec:method:opt}). The same
  configuration is used across all six LaMP tasks unless otherwise noted;
  per-task differences are limited to dataset paths and the length
  threshold $T$ (\cref{tab:ablation-length}).

  \vspace{0.5em}
  \noindent\textbf{GRPO (profile generator $\pi_\theta$).}
  The policy is optimized with TRL's GRPO trainer in bfloat16 with
  gradient checkpointing.
  \begin{itemize}[leftmargin=*,itemsep=1pt,topsep=2pt]
    \item Epochs: $2$
    \item Per-device train batch size: $4$
    \item Gradient accumulation steps: $20$ (effective batch
          $4 \times 20 \times 4 = 320$ over 4 training GPUs)
    \item Learning rate: $1.0\!\times\!10^{-5}$, warmup ratio $0.05$
    \item KL coefficient $\beta$: $0.01$
    \item Group size (rollouts per prompt) $G$: $8$
    \item Max completion length: $500$ tokens
    \item Save strategy: every $50$ steps, keep last $10$ checkpoints
  \end{itemize}
  The bidirectional reward is computed against $K{=}8$ in-batch negatives
  and averaged over $M{=}2$ random presentation orders. For BUMP+,
  $J_{\mathrm{hard}}{=}4$ of the $K{=}8$ negatives are hard-mined with
  BGE; the rest are sampled uniformly in-batch.

  \vspace{0.5em}
  \noindent\textbf{LoRA-SFT (downstream backbone).}
  Once $\pi_\theta$ is frozen and each user's profile is cached, we
  fine-tune a separate \textsc{Qwen3-4B-Instruct} backbone with a LoRA
  adapter on the LaMP $(\texttt{input}, \texttt{gold})$ pairs, conditioned
  on the cached profile via the template of \cref{fig:sft-input-template}.
  \begin{itemize}[leftmargin=*,itemsep=1pt,topsep=2pt]
    \item Epochs: up to $10$ with early stopping (patience $5$,
          threshold $0$) on dev marginalized accuracy
    \item Per-device train / eval batch size: $4\,/\,4$
    \item Gradient accumulation steps: $4$ (effective batch
          $4 \times 4 \times 8 = 128$ over 8 GPUs)
    \item Learning rate: $1.0\!\times\!10^{-5}$, $10$ warmup steps
    \item Max sequence length: $4200$ tokens
    \item Evaluation / save strategy: every $50$ steps, keep last $3$
          checkpoints
    \item Precision / memory: bfloat16, gradient checkpointing,
          $16$ data-loader workers with pinned memory
  \end{itemize}
  LoRA configuration: rank $r{=}32$, $\alpha{=}32$, dropout $0.05$,
  applied to all linear projections (\texttt{target\_modules=all-linear}),
  task type causal LM. We disable \texttt{load\_best\_model\_at\_end}
  under DDP because PEFT's adapter reload places every rank's adapter on
  \texttt{cuda:0}, which OOMs on an 8-GPU run; instead, we select the
  best checkpoint post-hoc using dev metrics.

  \vspace{0.5em}
  \noindent\textbf{Shared across stages.}
  Both stages use the same backbone family (\textsc{Qwen3-4B-Instruct-2507})
  and the same dev-set early-stopping discipline (10\% random split of
  training users; \cref{app:data}). The frozen judge $J$ is served via
  vLLM with guided-regex decoding on a dedicated pair of GPUs and is
  never updated. All reported numbers are averaged over $3$ random seeds
  sharing these settings.

\end{document}